\documentclass{article}

\usepackage[english]{babel}
\usepackage{subcaption}
\usepackage{pgffor}
\usepackage{authblk}
\date{} 
\providecommand{\keywords}[1]{\textbf{Keywords:} #1}
\usepackage[top=1in, bottom=1in, left=1in, right=1in]{geometry} 
\usepackage{amsmath,amssymb}

\usepackage{graphicx}
\usepackage[hidelinks]{hyperref}
\usepackage[nopatch=footnote]{microtype}
\usepackage{setspace}
\usepackage[square,numbers,sort&compress]{natbib}
\usepackage{xcolor}

\usepackage{tikz}
\usepackage[percent]{overpic}

\usepackage{doi}

\title{InSituRes: A Physics-Informed Same-Grid Model for Enhanced Dynamic X-ray Micro-CT Reconstructions}

\author[1]{Qinyi Tian}
\author[2]{Andrea Bisciotti}
\author[1]{Soniya Tiwari}
\author[3]{Spence Cox}

\author[1,*]{Laura E. Dalton} 

\affil[1]{Department of Civil \& Environmental Engineering, Duke University, Durham, North Carolina, 27708, USA}
\affil[2]{Department of Physics and Earth Sciences, University of Ferrara, Via Saragat 1, Ferrara 44141, Italy}
\affil[3]{Department of Mechanical Engineering \& Materials Science, Duke University, Durham, North Carolina, 27708, USA}

\affil[*]{Corresponding author email: laura.dalton@duke.edu}

\begin{document}

\maketitle
\doublespacing
\vspace{-13mm}
\begin{abstract}
X-ray micro-computed tomography (micro-CT) provides non-destructive three-dimensional (3D) imaging of porous material microstructures. \textit{In situ} experiments, including mechanical loading and reactive transport, increasingly require dynamic four-dimensional (4D) imaging with volumes repeatedly acquired during experiments. However, rapid acquisition typically requires fewer projections, shorter exposures, or reduced fields of view, producing reconstructions with noise, blur, and artifacts that obscure pores, microcracks, and interfaces. To address this challenge, this study introduces InSituRes, a physics-informed same-grid volumetric enhancement framework for fast dynamic X-ray micro-CT imaging of temporally evolving materials. InSituRes maps fast-acquisition volumes to higher-quality long-acquisition reconstructions using paired scans of the same specimens. The model integrates 3D convolutional feature extraction with slice-wise transformer attention to capture local and broader in-plane context. A learnable forward degradation model approximates rapid acquisition effects, including spatial blurring, intensity scaling differences, and signal-dependent noise. During training, reconstructed volumes should match high-quality reference scans and reproduce observed fast acquisition data after propagation through the forward model, imposing a physics-guided consistency constraint. Experiments on unseen micro-CT datasets demonstrate improved reconstruction fidelity and enhanced visibility of fine microstructural features relative to conventional interpolation and learning-based enhancement approaches. The framework supports quantitative interpretation of fast 4D X-ray micro-CT scans of evolving materials.

\end{abstract}


\newpage
\doublespacing
\section{Introduction}
X-ray micro-computed tomography (X-ray micro-CT) enables non-destructive three-dimensional (3D) imaging of the internal microstructures of opaque materials \cite{article1226, PAK2023100017, Katsamenis2019}. By resolving internal structures at micrometer scale spatial resolution, X-ray micro-CT has become an important tool for quantifying pore connectivity, manufacturing defects, fracture networks, and material phase distributions \cite{cnudde2013high, duPlessis2017MicroCTGuideline, bultreys2016microct}. These capabilities are particularly valuable for \textit{in situ} experiments, where the goal is to observe internal structural evolution under representative environmental or mechanical conditions rather than only comparing specimens before and after loading, chemical-induced reactions, or multiphase transport \cite{min9030183, ma18163895, FERDOUSH2023123565, stock2008microcomputed, maire2014quantitative, wildenschild2013xray, Waldner2024}. Newer X-ray micro-CT systems incorporate hardware and acquisition features that facilitate dynamic imaging by enabling rapid projection acquisition during in situ experiments, producing time-resolved, four-dimensional (4D) datasets \cite{FERDOUSH2023123565, Waldner2024, renard2017dynamic, Nwankwo2026, hall2010insitu, Klos2025, Dewanckele2020}. However, capturing evolving material states at high temporal resolution typically requires acquisition trade offs, including reduced projection counts over the $360^\circ$ rotation, shorter exposure times, increased detector readout rates, or reduced field of view \cite{Lin2009, Sombke2015, duPlessis2017MicroCTGuideline, bultreys2016microct, renard2017dynamic, Dewanckele2020}, which can increase noise, spatial blur, streaking artifacts, partial volume uncertainty, and reconstruction errors that obscure fine microstructural features such as small pores, microcracks, and thin particle boundaries \cite{maire2014quantitative, Usui2024, WINDOWSYULE20253, LIU2026103831}.

Most image enhancement methods are designed to either suppress degradation in existing images or to infer higher resolution images by enlarging the pixel or voxel grid \cite{9044873, LEPCHA2023230, SU2025110935}. Dynamic X-ray micro-CT poses a different problem because the acquisition geometry and reconstruction pipeline already fix the reconstructed voxel grid and field of view. Enhancement must therefore recover fine structural contrast on the same voxel grid while remaining consistent with the measured fast scan observation. Several approaches have been developed to improve degraded tomographic images, but most only partially address the challenges of dynamic X-ray micro-CT. Conventional filtering and denoising can suppress noise, but often smooth small pores, cracks, and interfaces \cite{maire2014quantitative}. Iterative reconstruction and compressed sensing methods improve sparse view reconstructions by imposing regularization constraints or sparsity priors \cite{wang2016perspective}; for example, Adaptive Steepest Descent onto Convex Sets (ASD-POCS) has been applied to sparse-view and low-dose reconstruction under limited-data conditions \cite{Sidky2008ASDPOCS, Champley2024, DALTON2022}. Super-resolution and learning-based enhancement have been explored across X-ray imaging, including tomographic super-resolution \cite{Dreier2021SuperResolutionTomography}, medical X-ray enhancement and low-dose chest X-ray recovery \cite{Shimizu2015SuperResolutionXray, Sano2017MedicalSuperResolution, Xu2020LowDoseCXR, Prasath2025DTRSN}, and geological material analysis \cite{Alqahtani2022SuperResolvedSegmentation, Omori2023RockCTSuperResolution}. These studies demonstrate the broad relevance of super-resolution for X-ray-based imaging, but most address static image enhancement, diagnostic restoration, segmentation support, or post-acquisition resolution improvement rather than dynamic X-ray micro-CT scan quality enhancement.

Same-grid enhancement improves the fast acquisition reconstruction on the original voxel grid, rather than producing an enlarged or resampled volume. This requirement is critical in dynamic experiments because visually plausible but physically inconsistent restoration can introduce artificial features and bias the interpretation of an evolving microstructure. While learning-based methods provide a powerful route for enhancing degraded X-ray micro-CT volumes \cite{https://doi.org/10.1029/2021WR031814, Ugolkov2025, UGOLKOV2025106018}, including convolutional neural networks for low-dose restoration, sparse-view reconstruction, and super-resolution in tomographic imaging \cite{Fok2024}, as well as X-ray tomography applications in digital core analysis and defect quantification \cite{Nwankwo2026, Klos2025}, many of these methods primarily learn statistical mappings between degraded and high-quality image domains and do not explicitly enforce consistency with the measured fast X-ray micro-CT observation \cite{f7lw-t7d7}. Thus, despite extensive prior work on X-ray image enhancement and super-resolution, few models specifically address dynamic X-ray micro-CT enhancement under the combined constraints of rapid acquisition, evolving material state, fixed voxel geometry, and observation consistency.

This study introduces InSituRes, a physics-informed same-grid volumetric enhancement framework for fast acquisition X-ray micro-CT scans, with particular focus on dynamic \textit{in situ} 4D imaging. The central hypothesis is that same-grid, physics-informed enhancement can improve fast acquisition X-ray micro-CT scans, particularly for \textit{in situ} 4D imaging, by recovering degraded microstructural contrast while maintaining consistency with the measured fast acquisition observation. InSituRes tests this hypothesis using paired fast and long acquisition X-ray micro-CT scans of the same specimens to train a reconstruction model coupled to a learnable forward degradation process. The framework is evaluated using composite concrete and recycled fine aggregate materials. These materials contain present unique features incuding multiscale pores, aggregate-hydrated cement interfaces, hydration products, microcracks, and partial volume boundaries that are difficult to resolve under fast acquisition conditions. By combining same-grid enhancement with fast acquisition observation consistency, InSituRes aims to improve quantitative interpretation of dynamic \textit{in situ} X-ray micro-CT while preserving the measurement basis of the physics of the original scan.


\section{Results}
\subsection{InSituRes framework for physics-informed micro-CT enhancement}
\begin{figure}[ht]
    \centering
    \includegraphics[width=1\linewidth]{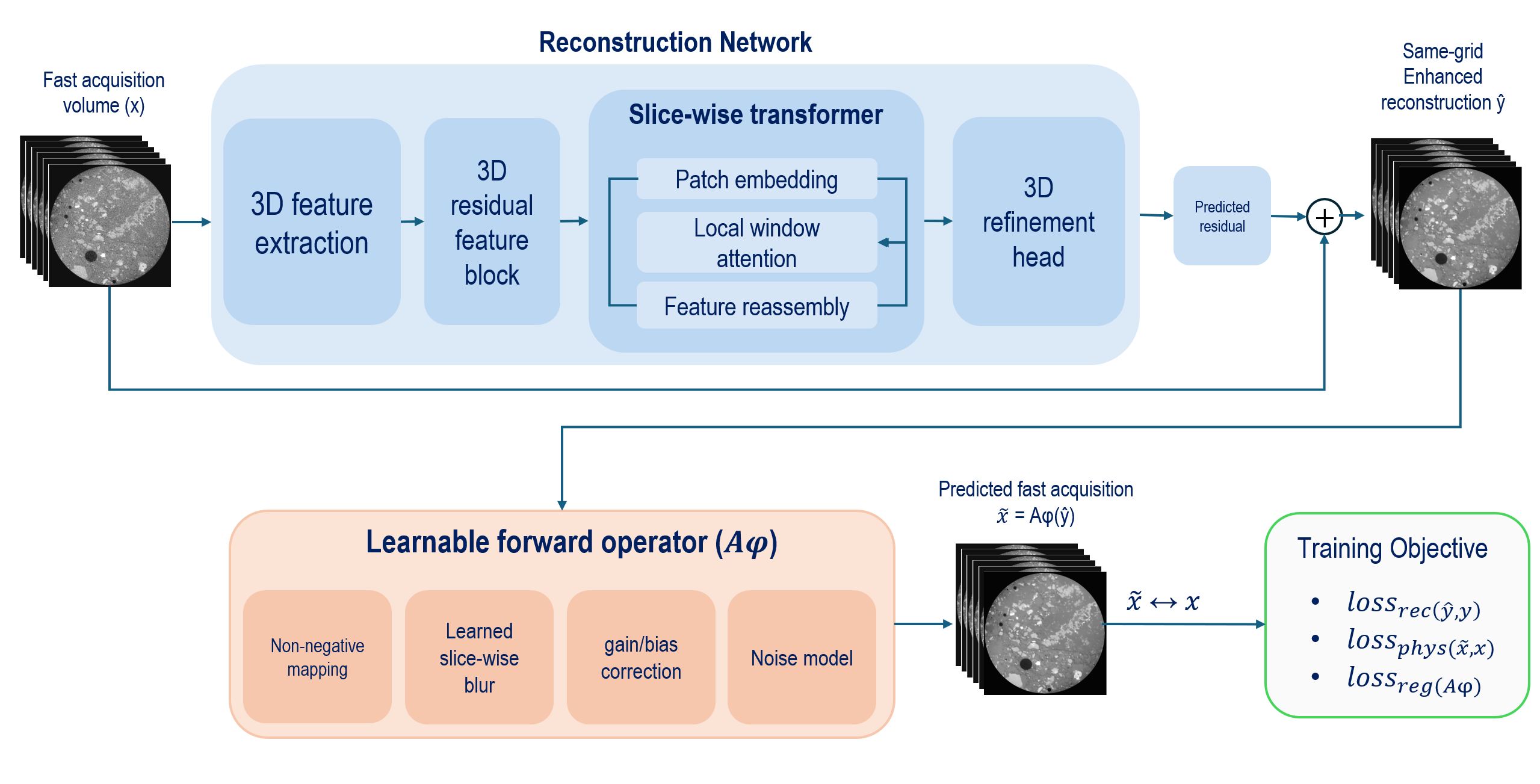}
   \caption{Architecture of InSituRes, a physics-informed framework for volumetric resolution enhancement with measurement consistent training.}
    \label{fig:insitures_arch}
\end{figure}

The reconstruction model, \textbf{InSituRes}, was designed to enhance degraded, fast acquisition X-ray micro-CT volumes while preserving the original voxel grid. The architecture of the proposed framework is illustrated in Figure~\ref{fig:insitures_arch}. Accordingly, the task was formulated as volumetric enhancement rather than spatial super-resolution, and the network predicts an improved reconstruction at the same spatial dimensions as the input. This distinction is important because the objective is not to infer missing voxels on a finer lattice, but to recover structural fidelity from fast scans that are corrupted by noise, blur, and reconstruction artifacts.

The framework contains two coupled components: a reconstruction network and a learnable forward degradation model. Although the slice-wise transformer improves spatial context modeling, it is not by itself the key innovation of the method. The scientific novelty of InSituRes is that the reconstruction network is not optimized as an unconstrained image restoration model. Instead, its output is passed through a learnable forward degradation model during training, so the enhancement is guided simultaneously by a high-quality reference scan and by consistency with the measured fast acquisition observation. This joint coupling between reconstruction and forward physics is what distinguishes the framework from standard supervised denoising and enhancement frameworks. The critical contribution of this formulation is the explicit closure of the reconstruction loop; the enhanced volume is not judged only by how closely it matches the long acquisition reference, but also by whether it can reproduce the measured fast acquisition scan after passing through a learnable degradation model. This coupling is the physical grounding in the proposed method.

\subsection{Resolution enhancement in dynamic X-ray micro-CT scans}
\begin{figure}[ht ]
\centering
\begin{subfigure}{0.325\textwidth}
\centering
\includegraphics[width=\linewidth]{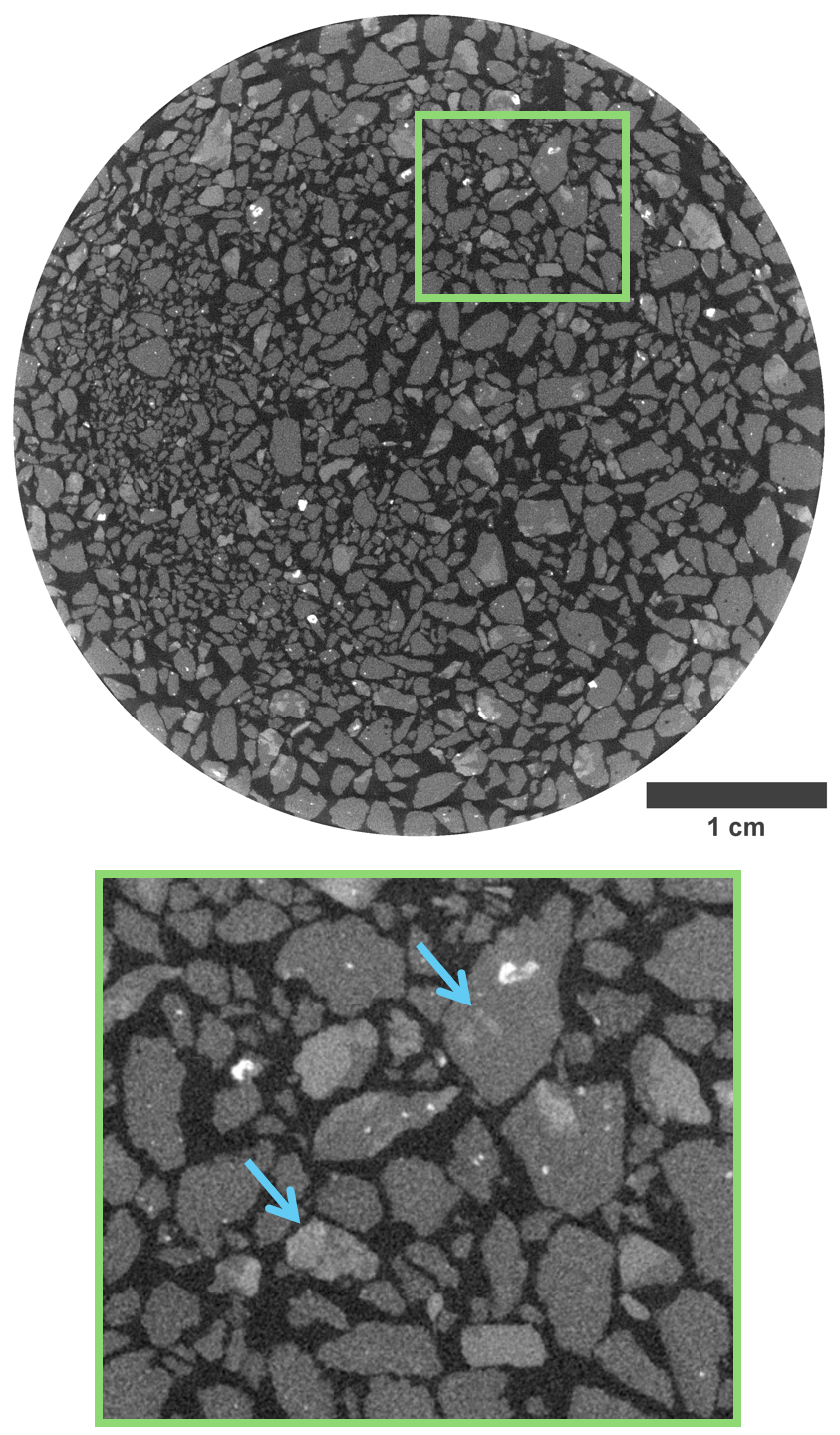}
\caption{}
\label{fig:reconstruction_comparison_a}
\end{subfigure}
\hfill
\begin{subfigure}{0.325\textwidth}
\centering
\includegraphics[width=\linewidth]{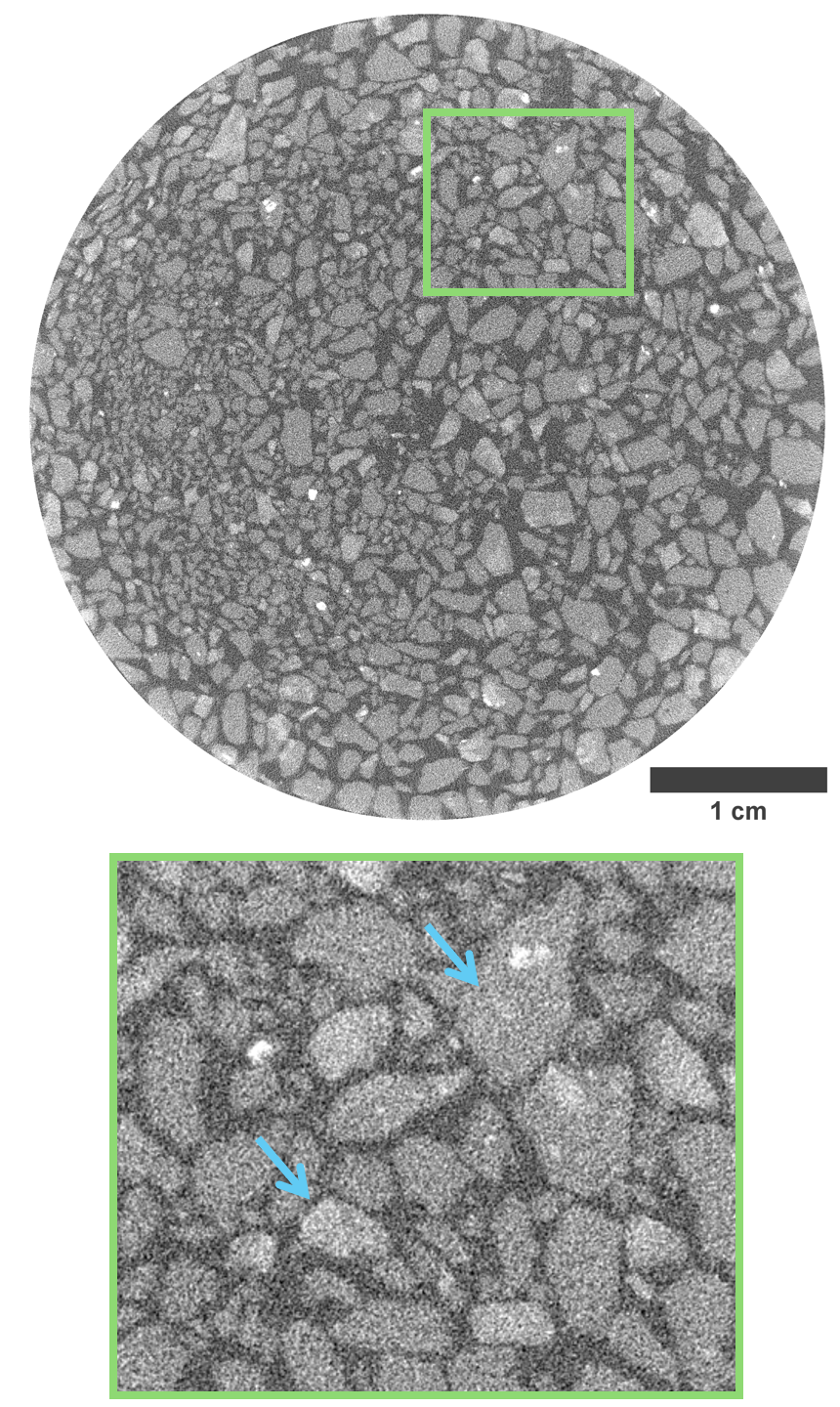}
\caption{}
\label{fig:reconstruction_comparison_b}
\end{subfigure}
\hfill
\begin{subfigure}{0.325\textwidth}
\centering
\includegraphics[width=\linewidth]{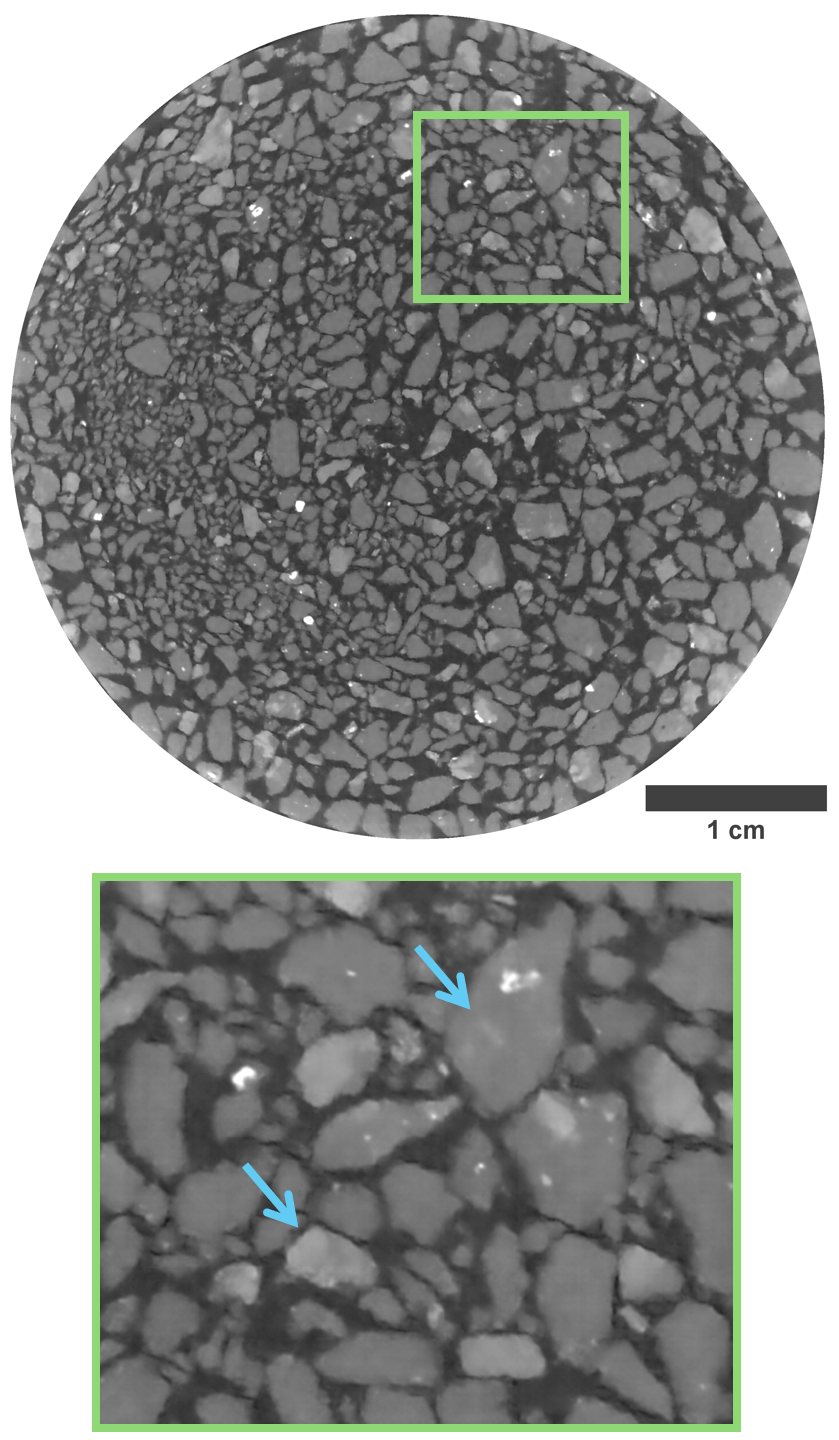}
\caption{}
\label{fig:reconstruction_comparison_c}
\end{subfigure}

\caption{Comparison of reconstruction quality for same spatial location slice from: (a) long acquisition reference scan, (b) fast acquisition scan, and (c) InSituRes enhanced. Green boxes mark the regions magnified in the second row, and blue arrows highlight local microstructural detail in the zoomed views.}
\label{fig:reconstruction_comparison}

\end{figure}

\begin{flushleft}
The InSituRes framework was applied to fast acquisition X-ray micro-CT scans of concrete materials acquired using a TESCAN UniTOM XL X-ray micro-CT system located in the Dalton Creativ Engineering Laboratory at Duke University \cite{TIAN2025100049} to evaluate its ability to enhance reconstruction quality under these conditions. Figure~\ref{fig:reconstruction_comparison} presents representative slices obtained from the standard long acquisition reference scan with an acquisition time of 35 minutes (Figure~\ref{fig:reconstruction_comparison_a}), the fast scan with an acquisition time of 4 minutes (Figure~\ref{fig:reconstruction_comparison_b}), and the corresponding reconstruction produced by InSituRes (Figure~\ref{fig:reconstruction_comparison_c}). The original fast acquisition reconstruction exhibits reduced contrast and blurred structural boundaries relative to the reference volume. In particular, pore edges and fine crack structures appear poorly defined due to noise and spatial smoothing introduced during rapid acquisition. In contrast, InSituRes produces enhanced reconstructions with improved structural clarity. The enhanced volume exhibits sharper pore-aggregate boundaries, improved visibility high density features, and reduced background noise compared to the original fast scan. Many structural features that are difficult to identify in the fast acquisition reconstruction become clearly visible after enhancement. The reconstructed volumes also maintain overall structural consistency with the reference scans, indicating that the enhancement process preserves the underlying microstructural organization of the specimen.
\end{flushleft}

\subsection{Quantitative reconstruction performance}

Quantitative performance was assessed by comparing both the fast acquisition input and the model-reconstructed output with the matched long acquisition ground truth (GT). Metrics were calculated across \(n = 91\) non-overlapping 3D slabs from the reconstructed volume (results showed in Table~\ref{tab:quantitative_reconstruction}. The InSituRes model output showed higher agreement with the long acquisition GT than the fast acquisition input, with Peak Signal-to-Noise Ratio (PSNR) increasing from 21.75 $\pm$ 0.23 decibels (dB) to 29.84 $\pm$ 0.30 dB and Structural Similaty Index Measure (SSIM) \cite{1284395} increasing from 0.3908 $\pm$ 0.0125 to 0.7397 $\pm$ 0.0112. This corresponded to improvements of 8.09 $\pm$ 0.19 dB in PSNR and 0.3489 $\pm$ 0.0092 in SSIM. Error-based metrics showed the same trend, with Mean Absolute Error (MAE) decreasing from 16.62 $\pm$ 0.36 to 5.45 $\pm$ 0.21 and Root Mean Square Error (RMSE) decreasing from 20.72 $\pm$ 0.53 to 8.17 $\pm$ 0.30.

\begin{figure}[htbp]
\centering
\includegraphics[width=0.95\textwidth]{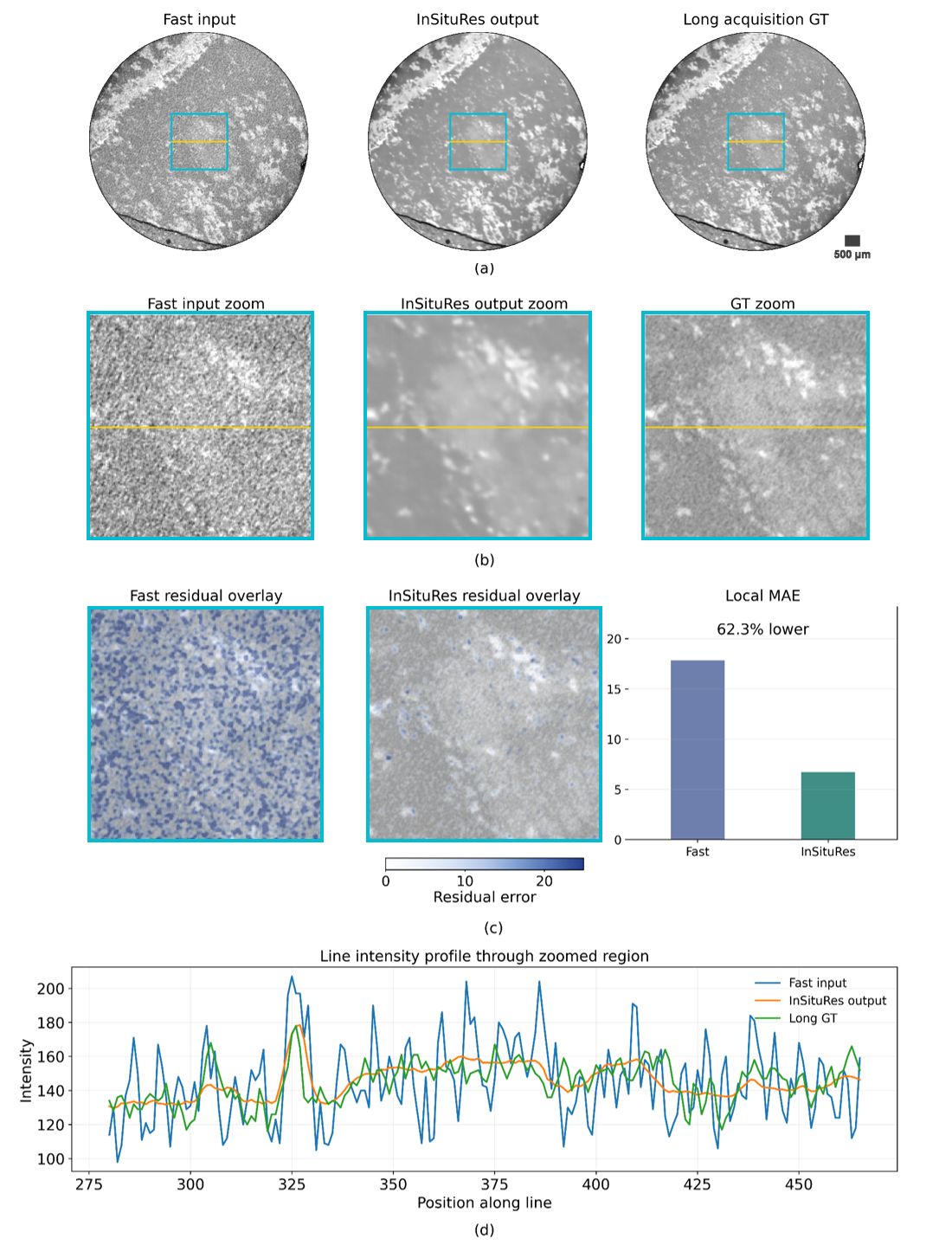}
\caption{
Representative reconstruction comparison.
(a) Full-slice views of the fast acquisition input, InSituRes reconstruction, and long acquisition GT.
(b) Zoomed regions corresponding to the blue boxes in (a).
(c) Local residual comparison using residual overlays and local MAE within the zoomed region.
(d) Line intensity profiles along the yellow lines in (a) and (b).
}
\label{fig:reconstruction_comparison-}
\end{figure}

Representative slice-level comparisons supported the quantitative improvements (Figure~\ref{fig:reconstruction_comparison-}a-d). Full-slice views provided spatial context for the fast acquisition input, InSituRes reconstruction, and long acquisition GT reference (Figure~\ref {fig:reconstruction_comparison-}a). The fast acquisition input exhibited a patchier appearance, consistent with higher acquisition noise and local intensity variation. In contrast, the InSituRes reconstruction appeared more spatially coherent while preserving the main structural patterns observed in the reference. This visual improvement was accompanied by lower residual error relative to the reference.

Zoomed regions further highlighted local structural and intensity differences between the fast acquisition input and the InSituRes reconstruction (Figure~\ref{fig:reconstruction_comparison-}b). Local residual overlays were calculated as the absolute difference between each image and the reference, with stronger residual signal indicating larger local deviation (Figure~\ref{fig:reconstruction_comparison-}c). The local mean absolute error (MAE) summary quantified this reduction by averaging the absolute intensity difference from the reference within the zoomed region. Line-intensity profiles through the same region showed closer agreement between the InSituRes reconstruction and the reference than between the fast acquisition input and the reference (Figure~\ref{fig:reconstruction_comparison-}d).

\begin{table}[htbp]
\centering
\renewcommand{\arraystretch}{1.5}
\caption{
Quantitative comparison between the fast acquisition input and the InSituRes reconstruction relative to the long acquisition reference. Values are reported as mean $\pm$ standard deviation across \(n = 91\) non-overlapping 3D slabs. Higher PSNR and SSIM indicate better reconstruction quality; lower MAE and RMSE indicate lower reconstruction error.
}
\label{tab:quantitative_reconstruction}
\begin{tabular}{p{5cm}cccc}
\hline
\textbf{Method} & \textbf{PSNR $\uparrow$} & \textbf{SSIM $\uparrow$} & \textbf{MAE $\downarrow$} & \textbf{RMSE $\downarrow$} \\
\hline
Fast acquisition input & 21.75 $\pm$ 0.23 & 0.3908 $\pm$ 0.0125 & 16.62 $\pm$ 0.36 & 20.72 $\pm$ 0.53 \\
InSituRes reconstruction & 29.84 $\pm$ 0.30 & 0.7397 $\pm$ 0.0112 & 5.45 $\pm$ 0.21 & 8.17 $\pm$ 0.30 \\
\hline
\end{tabular}
\end{table}

\subsection{Ablation analysis}
\begin{flushleft}
A controlled subset ablation analysis was performed to determine which components of InSituRes contributed to reconstruction performance. All trainable variants were trained from scratch using the same set of 1443 paired fast and long acquisition X-ray micro-CT slices and evaluated on the same held-out test data using PSNR and SSIM. The 1443 slice subset was split into training slices, validation slices, and test slices using an 80/10/10 split. The physics-informed loss was removed to test whether degradation consistent with the measured fast acquisition scan improved fidelity to the long acquisition reference. The transformer block was removed to test whether non-local contextual modeling contributed to spatially coherent reconstruction. Residual learning was removed to test whether predicting a correction to the fast acquisition input improved optimization compared with direct prediction of the reference image. A no-exponential moving average (no-EMA) evaluation was also performed to assess the effect of exponential moving average checkpoint weights during inference. The full InSituRes model achieved the best overall performance across the evaluated configurations (Table~\ref{tab:ablation}). Removing the physics-informed loss reduced reconstruction accuracy, indicating degradation consistency provided information beyond image-domain supervision alone. Removing the transformer block decreased performance, supporting the role of contextual feature modeling in recovering coherent microstructural patterns. The variant without residual learning showed lower fidelity, suggesting that residual formulation stabilized the reconstruction task by preserving information already present in the fast acquisition input while learning the missing correction. Evaluation without EMA weights further showed that checkpoint averaging improved inference robustness.

\begin{table}[htbp]
\centering
\renewcommand{\arraystretch}{1.35}
\caption{
Controlled ablation analysis using 1443 paired fast- and long-acquisition X-ray micro-CT slices. Values are reported as mean values on the held-out test data.
}
\label{tab:ablation}
\begin{tabular}{p{6.5cm}cc}
\hline
\textbf{Configuration} & \textbf{PSNR (dB) $\uparrow$} & \textbf{SSIM $\uparrow$} \\
\hline
Fast-acquisition input & $21.41$ & $0.3886$ \\
\textbf{Full InSituRes} & \textbf{29.56} & \textbf{0.7309} \\
InSituRes without physics-informed loss & $28.39$ & $0.6970$ \\
InSituRes without transformer block & $28.93$ & $0.7159$ \\
InSituRes without residual learning & $28.23$ & $0.7041$ \\
InSituRes without EMA inference & $28.56$ & $0.7060$ \\
\hline
\end{tabular}
\end{table}

\end{flushleft}

\subsection{Application to time resolved \textit{in situ} micro-CT imaging}
\begin{flushleft}
The performance of InSituRes is further assessed using time resolved \textit{in situ} micro-CT data acquired during uniaxial compression of a recycled fine aggregates specimen. This dataset was not included in training or validation and therefore provides an independent evaluation of model performance at unseen experimental conditions. The material system is composed of two physically meaningful phases, fine aggregate with residual hydrated cement phases and void space between the aggregates, which enables direct evaluation of both structural fidelity and quantitative phase recovery.

The dynamic acquisition consists of ten consecutive scans obtained during progressive loading. Each scan corresponds to a distinct compression state, capturing the evolution of pore structure as the void space between fine aggregates reduce and loading forces are redistributed between aggregate-aggregate contact. A single slice is extracted from each scan at a fixed spatial location, forming a temporally ordered sequence that allows direct tracking of microstructural changes. All slices are processed using InSituRes. Figure~\ref{fig:dynamic_results} presents the qualitative comparison. The first row in Figure~\ref{fig:dynamic_results} shows the raw fast acquisition reconstructions. These images exhibit pronounced noise, reduced contrast, and substantial overlap in intensity between pore space and aggregate. As a result, boundaries are poorly defined, and small pores are either obscured or indistinguishable from background fluctuations. The second row in Figure~\ref{fig:dynamic_results} shows the corresponding outputs from InSituRes. The enhanced images display sharper interfaces, improved contrast, and clearer delineation of pore geometry. The enhanced images show visually coherent structural evolution across the sequence, and no obvious discontinuous artifacts are observed in the representative slices examined.

\begin{figure}[!t]
\centering
\newcommand{\imgw}{0.17\textwidth}
\setlength{\tabcolsep}{1.5pt}
\renewcommand{\arraystretch}{0.9}
{\large increasing compression}

\vspace{0.5mm}

\begin{tikzpicture}
\draw[->, line width=0.9pt] (0,0) -- (0.92\textwidth,0);
\end{tikzpicture}

\vspace{1mm}

\begin{tabular}{c c c c}
 & {\large 1} & {\large 5} & {\large 10} \\
\rotatebox{90}{\large Raw}
&
\includegraphics[width=0.30\textwidth]{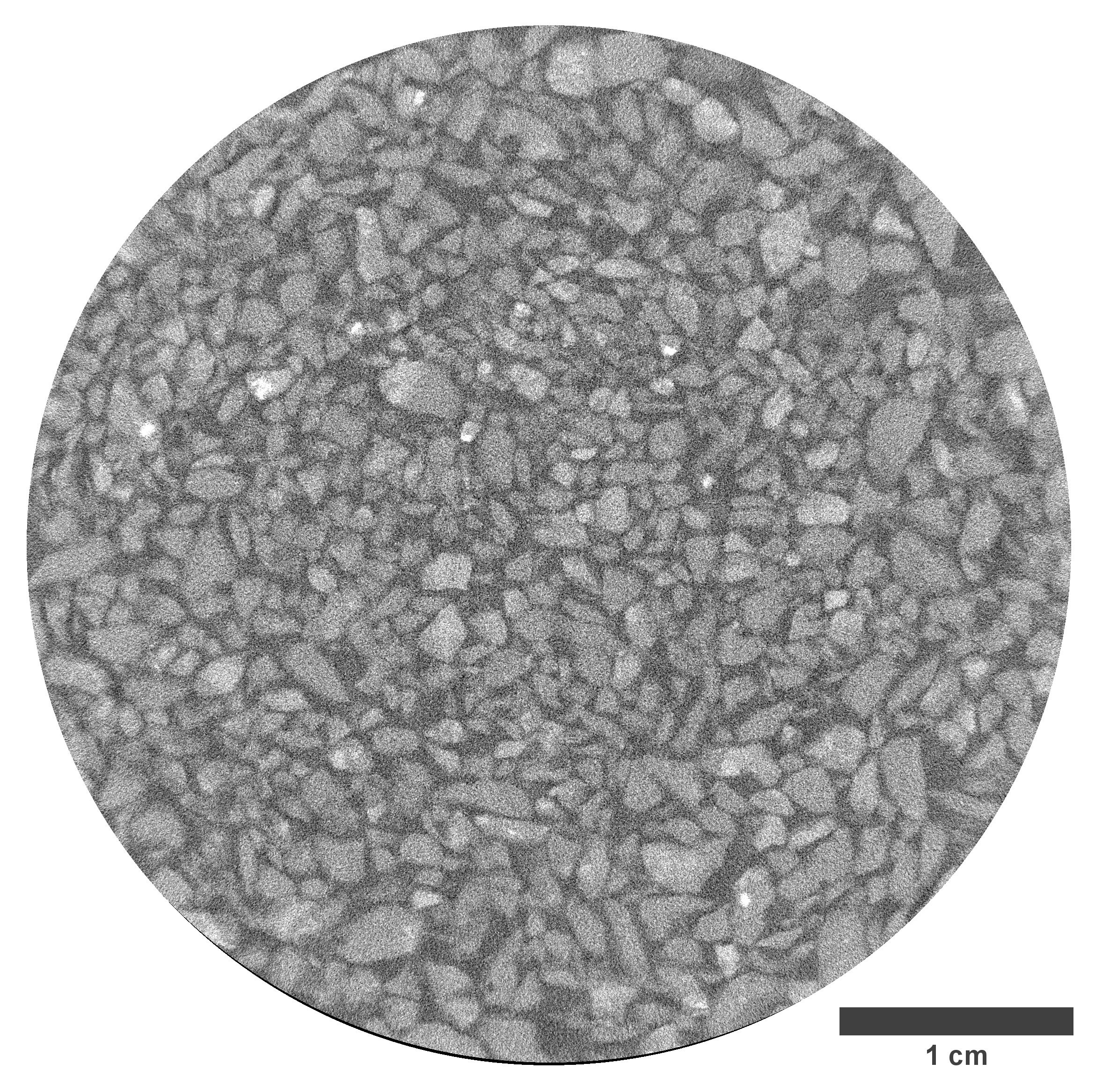}
&
\includegraphics[width=0.30\textwidth]{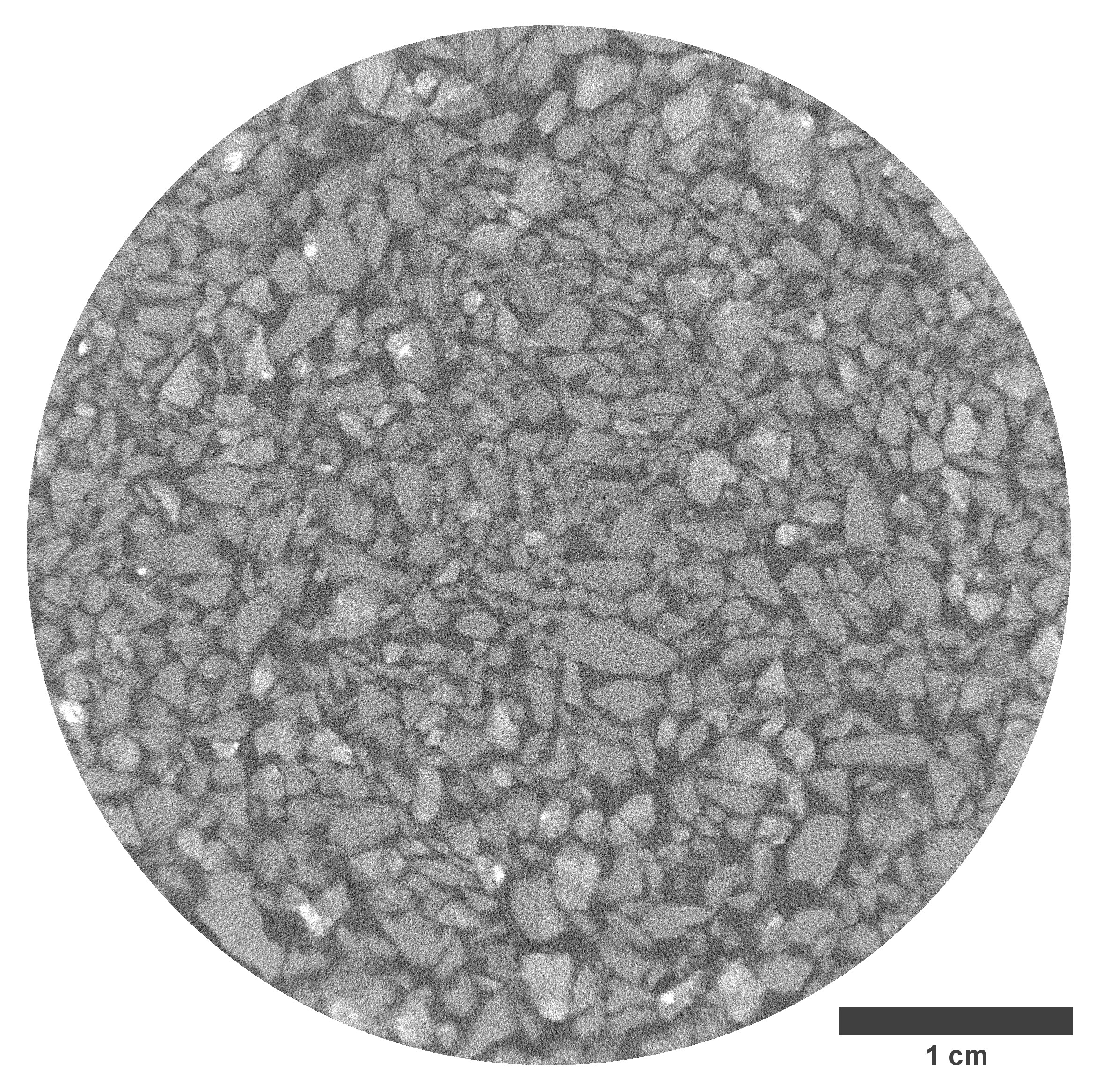}
&
\includegraphics[width=0.30\textwidth]{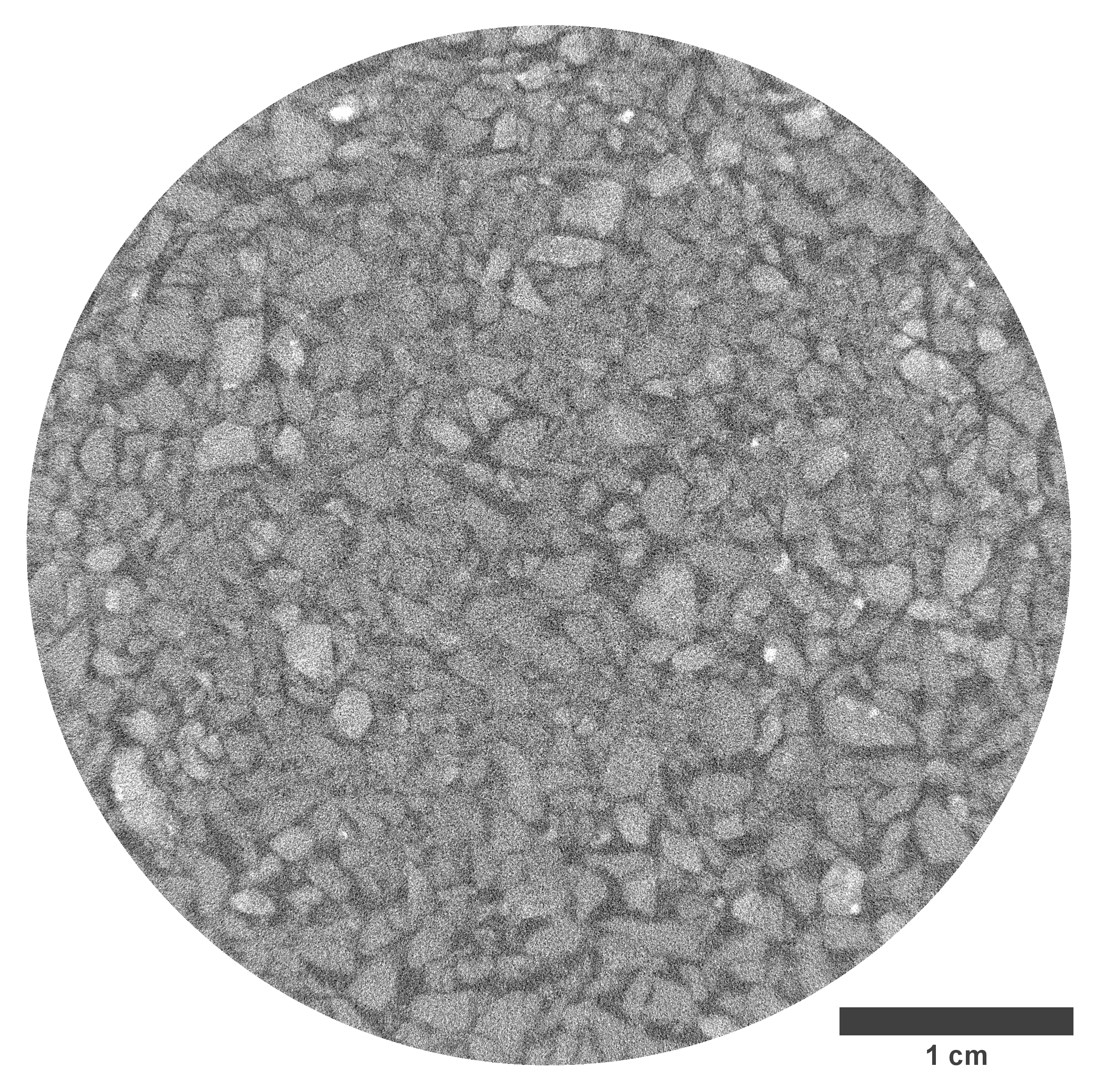}
\\[-0.5mm]

\rotatebox{90}{\large InSituRes}
&
\includegraphics[width=0.30\textwidth]{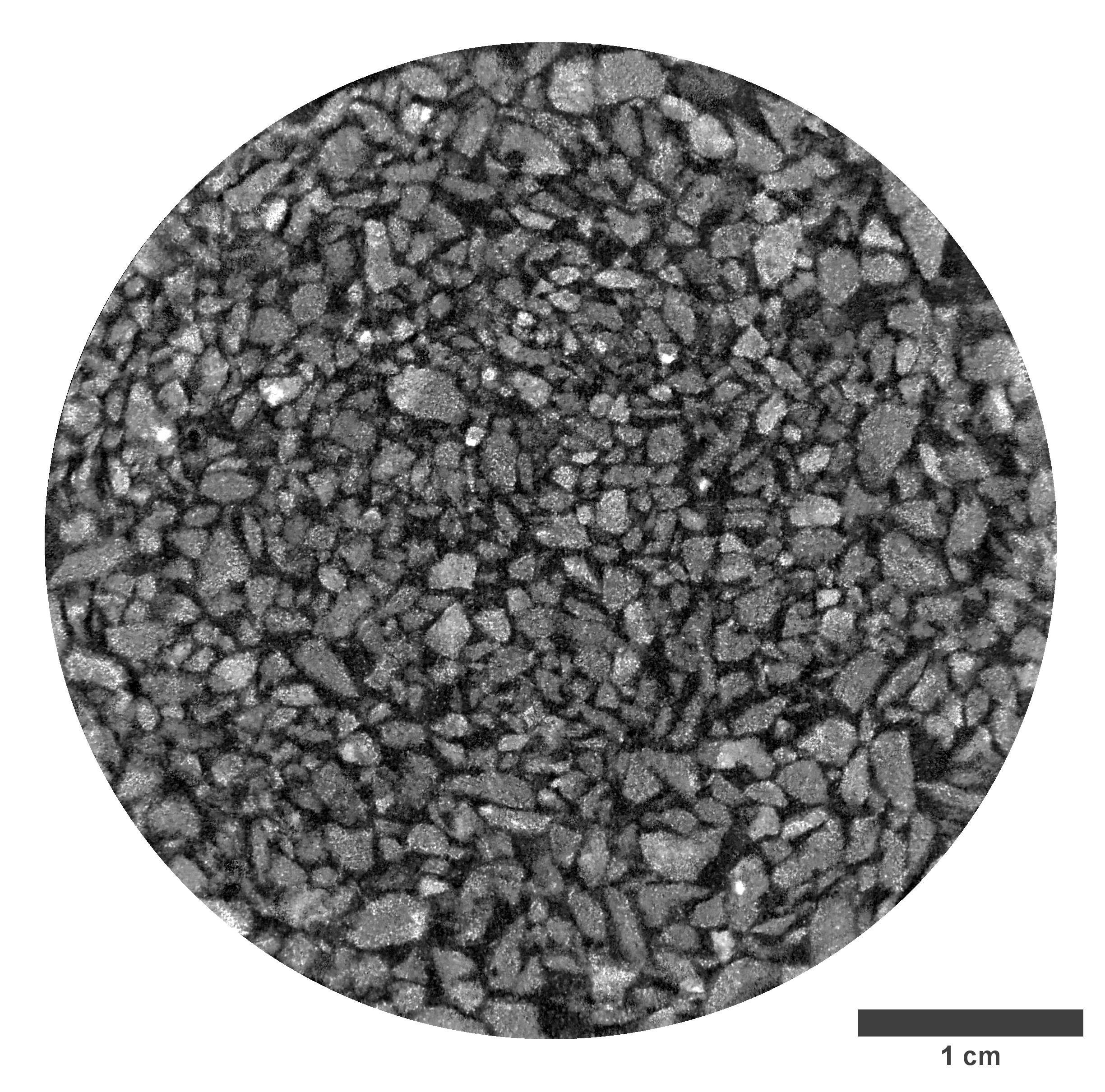}
&
\includegraphics[width=0.30\textwidth]{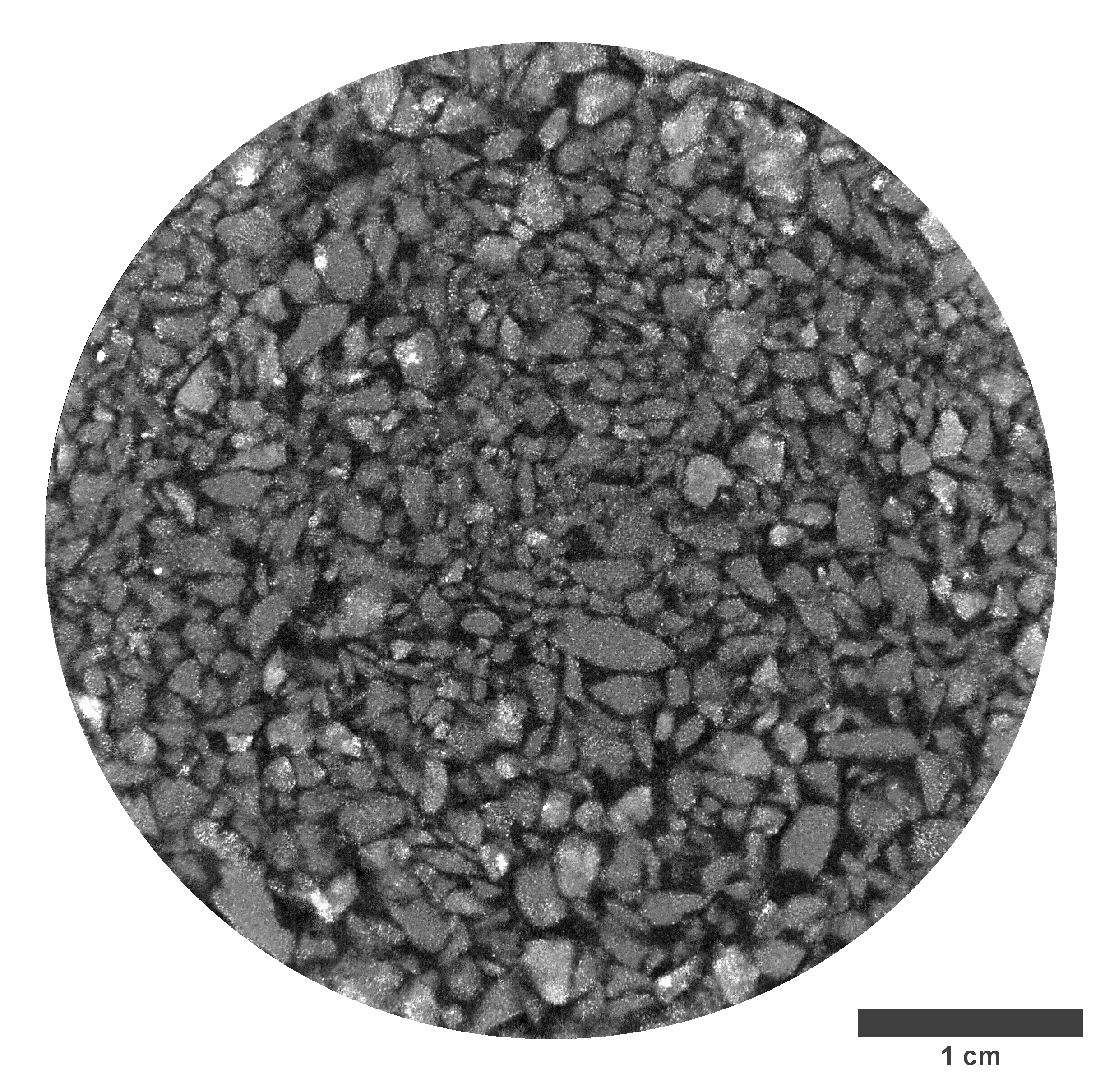}
&
\includegraphics[width=0.30\textwidth]{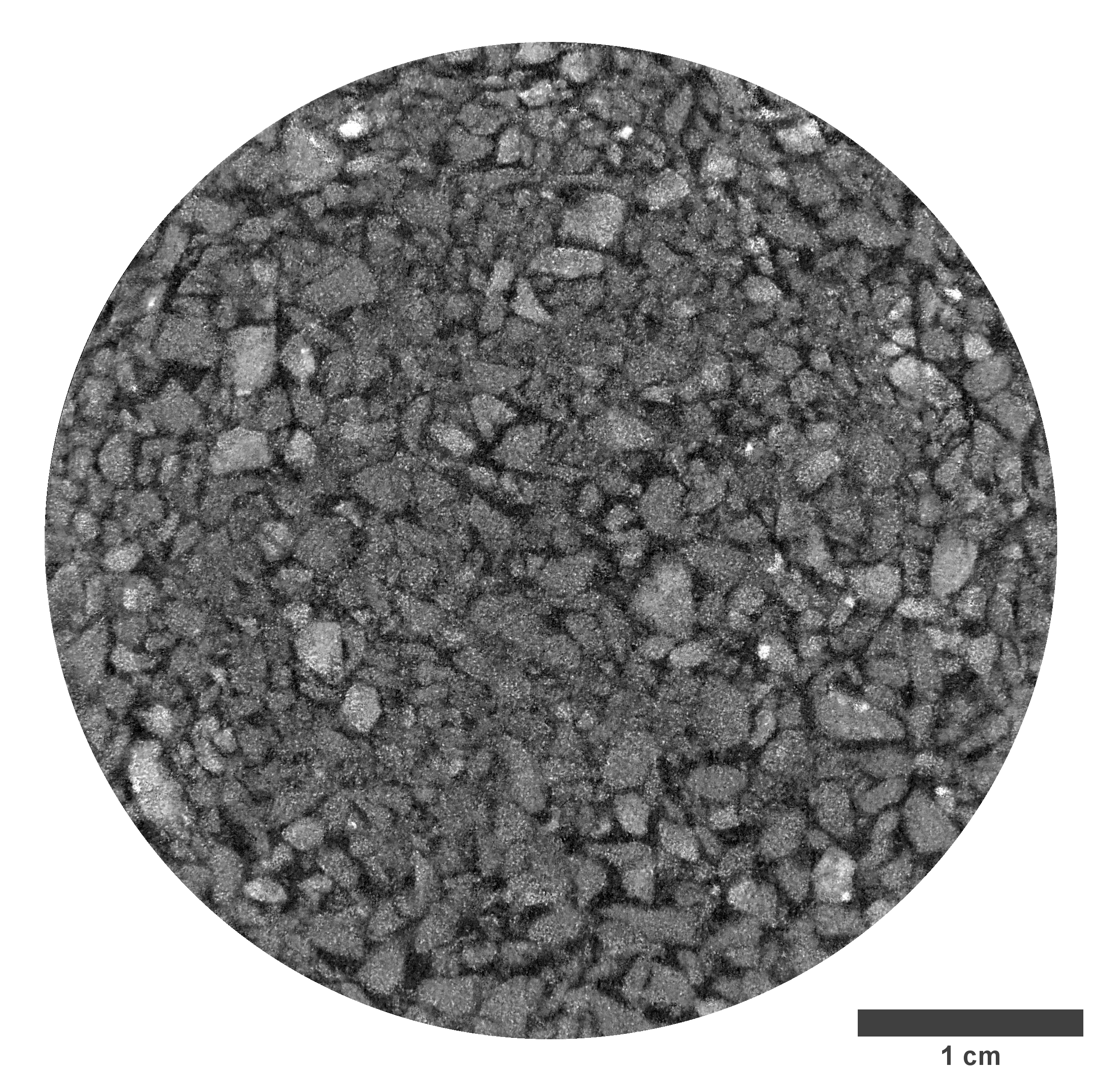}
\end{tabular}

\caption{Dynamic reconstruction of recycled aggregates during \textit{in situ} compression.
Representative early, middle, and late loading states are shown, corresponding to turns 1, 5, and 10 of the ten turn loading sequence. Each state pairs the raw fast acquisition reconstruction with the corresponding InSituRes output for direct comparison at the same deformation stage. InSituRes enhances aggregate boundaries and fine internal features in the representative loading states shown.}
\label{fig:dynamic_results}

\end{figure}

    
A central component of the evaluation is segmentation, which is used not as a visualization aid but as a quantitative probe of whether the reconstructed intensities correspond to physically meaningful phase separation. An unsupervised two-cluster k-means segmentation was applied to all datasets using identical parameters \cite{zis-MacQueen1967Some}. Given the binary nature of the material system, segmentation focuses on separating pore space from aggregate. Cluster assignments are made consistent across slices by ordering intensity centroids, ensuring that low-intensity regions correspond to pores and high-intensity regions correspond to aggregate throughout the entire dataset. Figure~\ref{fig:segmentation_results} illustrates this analysis for a representative case. The top row shows a slice from the first dynamic scan together with its segmentation. The bottom row shows the corresponding InSituRes output and its segmentation. In the raw reconstruction, segmentation is unstable due to noise and poor contrast, leading to fragmented pore regions and misclassification at phase boundaries. In contrast, segmentation of the enhanced image produces spatially coherent pore regions and well-defined aggregate boundaries. These results suggest that the enhanced images provide improved intensity separability between pore space and aggregate, which can support more consistent phase identification.

\begin{figure}[h]
    \centering
    \includegraphics[width=1\linewidth]{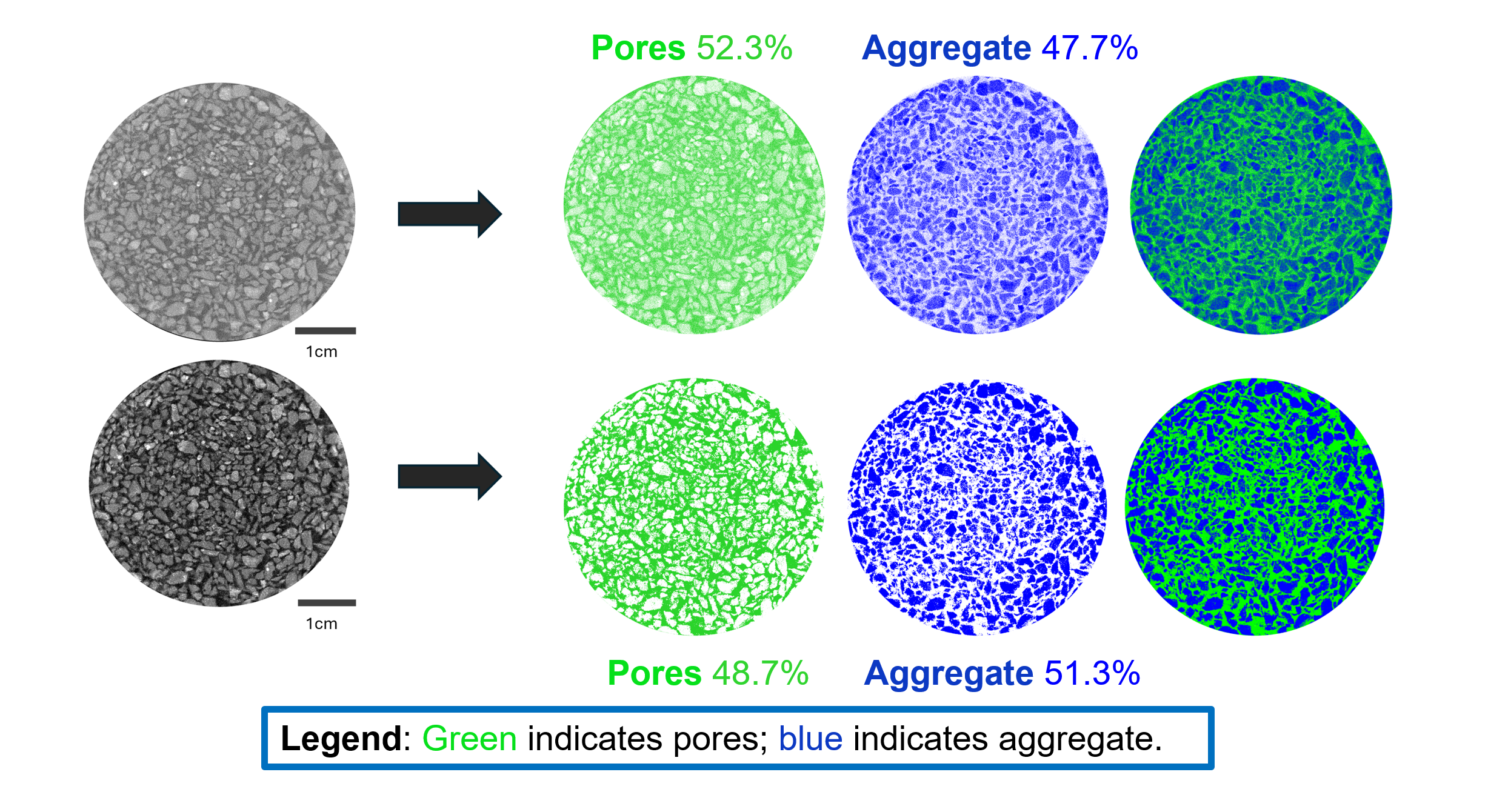}
   \caption{Segmentation comparison of raw and InSituRes-enhanced dynamic micro-CT data. 
Top: raw slice from the first dynamic scan with segmentation. 
Bottom: corresponding enhanced slice and segmentation. 
Pore space (green) and aggregate (blue) are more clearly separated after enhancement.}
    \label{fig:segmentation_results}
\end{figure}

To determine whether the improved segmentation quality translated into a quantitative phase measurement, porosity was measured for the representative slice shown in Figure~\ref{fig:reconstruction_comparison}, which contains the matched long acquisition reference scan, fast acquisition scan, and InSituRes-enhanced output at the same spatial location. Gaussian filtering, median filtering, bilateral filtering, and anisotropic diffusion were applied to the same fast acquisition slice. For each conventional filter family, the parameter setting that produced the lowest absolute porosity error relative to the long acquisition reference scan is reported in Table~\ref{tab:representative_slice_porosity_baselines}.

\begin{table}[h]
\centering
\renewcommand{\arraystretch}{1.35}
\caption{
Representative slice porosity comparison for the slice shown in Figure~\ref{fig:reconstruction_comparison}. The comparison includes the long acquisition reference scan, fast acquisition scan, conventional filtering baselines, and InSituRes-enhanced output. For each conventional filter, the parameter setting with the lowest absolute porosity error relative to the long acquisition reference scan is reported here in this table.
}
\label{tab:representative_slice_porosity_baselines}
\begin{tabular}{lcc}
\hline
\textbf{Method} & \textbf{Porosity (\%)} & \textbf{Absolute error (percentage points)} \\
\hline
Long acquisition reference scan & 48.34 & - \\
Fast acquisition scan & 55.70 & 7.36 \\
Gaussian filter & 49.10 & 0.77 \\
Median filter & 49.58 & 1.25 \\
Bilateral filter & 52.98 & 4.64 \\
Anisotropic diffusion & 53.19 & 4.85 \\
\textbf{InSituRes-enhanced output} & \textbf{48.55} & \textbf{0.21} \\
\hline
\end{tabular}
\end{table}

The fast acquisition scan overestimated porosity relative to the long acquisition reference scan. Conventional filtering reduced this error to varying degrees, with Gaussian filtering giving the lowest error among the filtering baselines. InSituRes produced the closest porosity estimate to the long acquisition reference scan, reducing the absolute porosity error from 7.36 to 0.21 percentage points.

To extend this analysis beyond individual slices, segmentation was applied to the full volumetric data. Phase fractions were computed for three cases: the long acquisition reference scan, the first dynamic fast acquisition scan, and the corresponding InSituRes-enhanced volume. The long acquisition reference scan was acquired under identical experimental conditions but with a longer acquisition time, providing higher signal-to-noise ratio and more reliable phase contrast. It therefore served as a higher quality reference for estimating the phase distribution. Full volume porosity measurements are summarized in Table~\ref{tab:dynamic_porosity_validation} and visualized in Figure~\ref{fig:porosity_comparison}. The fast acquisition scan overestimated the mean porosity relative to the long acquisition reference scan, reflecting bias introduced by noise and unresolved structure. In contrast, the InSituRes-enhanced volume reduced the absolute mean porosity error from 13.48 to 0.13 percentage points and reduced the slice-wise porosity MAE from 13.48 to 0.75 percentage points. These results indicate that InSituRes improves the consistency of porosity measurements with the long acquisition reference scan.

\begin{table}[htbp]
\centering
\renewcommand{\arraystretch}{1.45}
\caption{
Quantitative validation of dynamic \textit{in situ} micro-CT enhancement using full-volume porosity measurements across \(n = 882\) slices. Porosity was measured from the long acquisition reference scan, the first dynamic fast acquisition scan, and the corresponding InSituRes-enhanced volume. Absolute and relative errors were calculated relative to the long acquisition reference scan.
}
\label{tab:dynamic_porosity_validation}
\begin{tabular}{lccc}
\hline
\textbf{Metric} & \textbf{Long reference} & \textbf{Fast scan} & \textbf{InSituRes} \\
\hline
Mean porosity (\%) & 49.01 & 62.49 & 49.14 \\
Standard deviation (\%) & 1.11 & 0.67 & 1.31 \\
Mean porosity error (percentage points) & - & 13.48 & 0.13 \\
Relative mean porosity error (\%) &  - & 27.50 & 0.26 \\
Slice-wise MAE (percentage points) &  - & 13.48 & 0.75 \\
Slice-wise RMSE (percentage points) &  - & 13.51 & 1.06 \\
\hline
\end{tabular}
\end{table}

\begin{figure}[ht]
    \centering
    \includegraphics[width=1\linewidth]{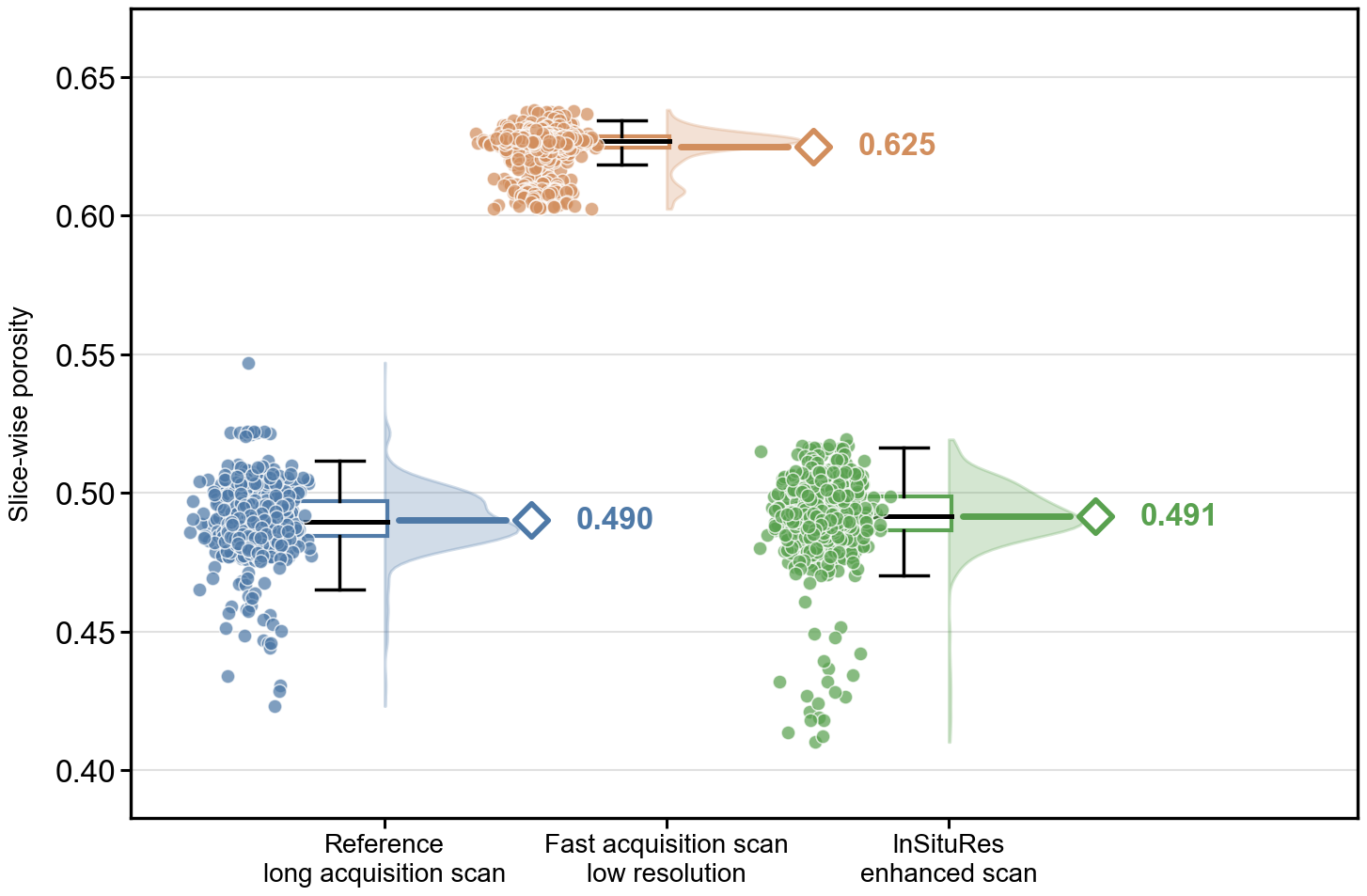}
\caption{Full volume comparison of slice wise porosity for the reference long acquisition scan, fast low resolution acquisition scan, and InSituRes enhanced reconstruction. Individual points denote porosity values measured from each slice, box plots show the median and interquartile range, and half violin plots indicate the overall distribution within each image stack. The InSituRes enhanced reconstruction more closely follows the reference distribution than the fast low resolution scan.}
    \label{fig:porosity_comparison}
\end{figure}

The combination of improved segmentation stability, reduced porosity error, and agreement with the long acquisition reference scan indicates that InSituRes enhances microstructural information in a manner consistent with the measured fast acquisition data. The application to an unseen recycled aggregate dataset further suggests that the method can generalize to related material types and imaging conditions. These results suggest that InSituRes can improve the analysis of time resolved \textit{in situ} experiments by reducing the gap between fast acquisition and quantitative interpretation.

\end{flushleft}

\subsection{Impact on microstructural feature visibility and analysis}
\begin{flushleft}

Improved reconstruction quality is particularly important for quantitative microstructural analysis of X-ray micro-CT data. Many experimental studies rely on segmentation and structural analysis of reconstructed volumes in order to quantify pore structure, crack development, and phase distribution within geomaterials and cement-based materials. Degradation in fast acquisition scans often reduces contrast between phases and obscures small structural features, which can significantly affect the reliability of segmentation and subsequent quantitative measurements.

To evaluate the impact of the proposed enhancement framework on downstream analysis, segmentation was performed on both the original fast acquisition scans and the enhanced volumes generated by InSituRes. 
Segmentation results qualitatively reflect the improved phase separability after enhancement. Segmentation performed on the original fast acquisition scan exhibits poorly defined pore boundaries and frequent merging of small pores with surrounding aggregate regions due to noise and reduced contrast. In contrast, segmentation performed on the enhanced reconstruction produced by InSituRes exhibits improved delineation of pore boundaries and more consistent phase separation. Small pores that are difficult to detect in the original fast scans become distinguishable after enhancement. This improves the reliability of pore identification within the reconstructed volumes.

Quantitative comparisons (Figure~\ref{fig:porosity_comparison}) further demonstrate the benefit of the enhancement for microstructural analysis. Metrics describing pore structure, including pore volume fraction and phase fraction, show improved agreement with values obtained from the long acquisition reference scans when computed from the enhanced volumes. These results indicate that the enhanced reconstructions more accurately preserve the underlying microstructural characteristics of the specimen. The improved segmentation quality obtained using InSituRes is particularly relevant for dynamic imaging experiments in which microstructural evolution must be tracked across successive time steps. Enhanced structural clarity facilitates more reliable identification of evolving features such as crack propagation paths and changes in pore distribution during mechanical loading. As a result, the proposed enhancement framework supports improved quantitative interpretation of time resolved micro-CT datasets.

Overall, these results demonstrate that the improvements provided by InSituRes extend beyond visual enhancement of reconstructed images. By improving phase separability, the framework enables more reliable quantitative analysis of pore structure and microstructural evolution in dynamic X-ray micro-CT experiments.

\end{flushleft}

\section{Discussion}
\begin{flushleft}
The results presented in this study demonstrate that InSituRes improves the spatial fidelity of fast acquisition X-ray micro-CT scans while preserving the original voxel grid required for time resolved comparison. Enhanced reconstructions exhibit clearer pore boundaries, improved crack visibility, and reduced noise relative to the original fast scans, with improved agreement against long acquisition reference scans. These improvements are important for dynamic \textit{in situ} experiments, where rapid volumetric acquisition is needed to capture evolving microstructural processes but reductions in projection number, exposure time, or related acquisition parameters can degrade reconstructed image quality. A key factor contributing to the performance of InSituRes is the integration of a physics-guided forward degradation model that constrains enhanced volumes to remain consistent with the measured fast acquisition data. By requiring enhanced reconstructions to reproduce the observed fast scan after propagation through the learned degradation model, InSituRes reduces the risk of generating visually plausible but physically unsupported structures. The ablation analysis supports this interpretation, as removing the forward model reduces reconstruction quality and suggests greater dependence on learned image priors, particularly under strong degradation.

The improved reconstructions have direct implications for downstream microstructural analysis. Quantitative interpretation of \textit{in situ} X-ray micro-CT datasets often depends on accurate identification and segmentation of pores, cracks, inclusions, fibers, grains, phase boundaries, and other internal structures. Degradation in fast scans can bias segmentation and propagate into derived quantities such as porosity, crack density, connectivity, tortuosity, permeability estimates, phase fractions, and damage metrics. By improving the visibility of fine structures while maintaining consistency with the measured fast acquisition data, InSituRes supports more reliable extraction of microstructural descriptors from time resolved scans. This capability is particularly relevant for experiments investigating fracture development, pore evolution, deformation, transport, and damage processes in porous, composite, granular, cement-based, and multiphase materials.

Several limitations remain. InSituRes requires paired fast acquisition and long acquisition scans during model training or fine-tuning, but not during deployment. Once trained, the pretrained model can enhance fast acquisition X-ray micro-CT volumes without requiring a corresponding long acquisition reference scan for each new experiment. Therefore, the main limitation is the extent to which the paired training data represent new imaging systems, material classes, and acquisition protocols. In addition, the forward degradation model approximates selected characteristics of fast acquisition rather than providing a complete physical representation of tomographic image formation. Future work should explore more detailed acquisition models, self-supervised or weakly supervised adaptation strategies, domain adaptation across materials and scanners, and validation against downstream physical measurements rather than image similarity alone. Overall, InSituRes provides a practical route for applying pretrained same-grid enhancement models to fast dynamic \textit{in situ} X-ray micro-CT scans, supporting quantitative analysis when conventional high-quality scanning is too slow to capture the evolving material state.

\end{flushleft}

\section{Materials and Methods}
\subsection{Overview}
\begin{flushleft}
Time resolved X-ray micro-CT provides a powerful tool for investigating the internal evolution of porous materials under dynamic conditions, including mechanical loading, multiphase transport, and microstructural chemical rearrangement. In these experiments, repeated scans are acquired as the specimen undergoes microstructural changes, producing a 4D dataset comprising 3D volumes that evolve over time. Achieving sufficient temporal resolution during \textit{in situ} experiments requires rapid acquisition on the order of minutes to seconds, often with reduced projection counts and shortened exposure times. Although these strategies enable faster scanning, they introduce substantial degradation in the reconstructed image volumes, including increased noise, reconstruction artifacts, and spatial blurring. These degradations reduce the visibility of critical microstructural features such as pores, hydrated phase boundaries, and microcracks, which are essential for interpreting deformation mechanisms in geomaterials.

In this work, the compromise of high spatial and high temporal resolution is addressed using a learning-based volumetric enhancement framework, InSituRes, designed to improve the quality of fast acquisition micro-CT scans without requiring modifications to the scanning hardware or acquisition protocol. The approach leverages paired training data consisting of fast acquisition scans and corresponding high-quality reference scans acquired using longer exposure times or higher projection counts. During training, the deep neural network learned a mapping scheme that transforms degraded fast scans into enhanced volumes that approximate the structural fidelity of the reference scans. Once trained, the model was applied to fast acquisition scans acquired at the temporal resolution required for 4D imaging. The physics-informed component was included to encourage consistency with measured fast acquisition observations and reduce the likelihood of generating structures incompatible with the imaging process.


\subsection{Materials}
\subsubsection{Model training material}
\begin{flushleft}
X-ray micro-CT scans of small pieces of concrete obtained from 4 inch diameter by 8 inch length (100 mm diameter by 200 mm length) concrete specimen after mechanically breaking the cylinder were used to train the model. The concrete cylinder was prepared using Type I/II Ordinary Portland Cement (OPC) with a water-to-cement ratio of 0.45 and fine and coarse aggregates were supplied by The Sunrock Group, LLC from the Butner Quarry in North Carolina, USA. The fine aggregate was manufactured sand (Plant ID BCS-CS) and the coarse aggregate was No. 67 gravel (Plant ID B67-67).
\end{flushleft}

\subsubsection{Independent external validation material}
\begin{flushleft}

A separate micro-CT dataset was used for independent external validation. The dataset consisted of scans of recycled aggregate fines produced in the laboratory from the comminution of concrete cylinders following 28 days of curing. The recycled fines included all crushed materials passing a No. 4 sieve ($> 4.75$ mm) and were primarily composed of natural olivine sand and basalt aggregates surrounded by adhered hydrated cement paste (Type I OPC). Samples were analyzed either in their untreated state or after accelerated CO$_2$ treatment under semi-dry conditions, aimed at densifying and strengthening the residual cement paste microstructure. Both long acquisition scans before mechanical detachment and fast dynamic scan slices acquired during in situ mechanical compression were collected using a TESCAN UniTOM XL X-ray micro-CT system equipped with a Deben CT5000-RT in situ 5 kN load cell. For this work, both long acquisition scans and dynamic scan slices from the \textit{in situ} compression experiments were used for model validation, enabling evaluation of the influence of mechanical loading on the microstructure of the different phases. These external scans were kept completely separate from the datasets used for model training, validation, and internal testing and were included solely to evaluate the capability of the trained model to generalize to newly acquired independent micro-CT data.
   
\end{flushleft}

\subsection{X-ray micro-CT imaging data acquisition}
\subsubsection{X-ray Micro-CT scanner}
All X-ray micro-CT data were acquired using a TESCAN UniTOM XL X-ray micro-CT system located in the Dalton Creativ Engineering Laboratory at Duke University. Unlike conventional laboratory micro-CT platforms that are optimized primarily for static imaging, this system supports continuous dynamic acquisition through a slip ring technology, enabling uninterrupted rotation during scanning while preventing cable wrap. The system also supports rapid tomography, with scan times reported to be as short as 10 seconds for a full 360 \textdegree rotation under optimized conditions \cite{tescan_unitom_xl}. These features distinguish the platform from conventional step and shoot micro-CT systems and make it well suited for dynamic imaging applications. The system has also been applied in previous micro-CT studies for high resolution imaging and image based segmentation tasks, such as pore-network characterization in sandstone samples \cite{TIAN2025100049}.

\subsubsection{Model training dataset}
X-ray micro-CT data used for training InSituRes were acquired from a section extracted from a $4 \times 8$ inches fractured concrete cylindrical sample with water to cement ratio of 0.45. 
Two micro-CT scans were collected using a TESCAN UniTOM XL scanner under identical acquisition conditions, including tube voltage of 120 kV, beam power of 15 W, 1,597 projections, voxel size of 10 $\mu$m, rotation range from 0° to 360°, one frame averaging, original pixel size of 0.15, and identical initial and final rotation-stage positions. The only acquisition parameter intentionally varied between the two scans was the exposure time. 
The fast scan was acquired using an exposure time of 90 milliseconds (ms), whereas the long exposure scan was acquired using an exposure time of 1200 ms. These exposure times were selected based on the detector transmission response of the cement-based specimens, with the objective of maximizing the detected signal within the usable dynamic range of the X-ray micro-CT system while avoiding detector saturation. The scanner has an upper practical intensity limit of 45,000 photon counts; therefore, the acquisition settings were adjusted to maximize the transmitted signal without exceeding the recommended range. Histogram inspection during acquisition confirmed that the detector response remained within the usable range, with maximum values of approximately 44,000. The 90 ms exposure, therefore, represented a rapid low exposure scan with undersampled photon statistics, whereas the 1200 ms exposure represented a long exposure reference scan with improved signal-to-noise ratio. Both scans were reconstructed using the built-in TESCAN Panthera\textsuperscript{TM}\cite{10.1093/mictod/qaae056} reconstruction software using an identical reconstruction pipeline for the fast- and long-exposure scans and producing volumetric datasets with the same number of cross-sectional slices. Because the scan geometry, rotation stage positioning, and reconstruction procedure were held constant, the reconstructed slices were spatially registered between the two volumes. This resulted in paired X-ray micro-CT datasets consisting of a low exposure fast-scan volume and a high exposure long scan reference volume. One paired fast scan and long scan is shown in Figure~\ref{fig:sample_visualization} as a visual example.

\begin{figure}[ht]
\centering
\includegraphics[width=1\textwidth]{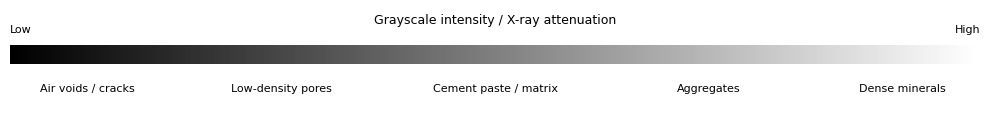}
\vspace{2mm}
\begin{minipage}{0.49\textwidth}
\centering
\includegraphics[width=\linewidth]{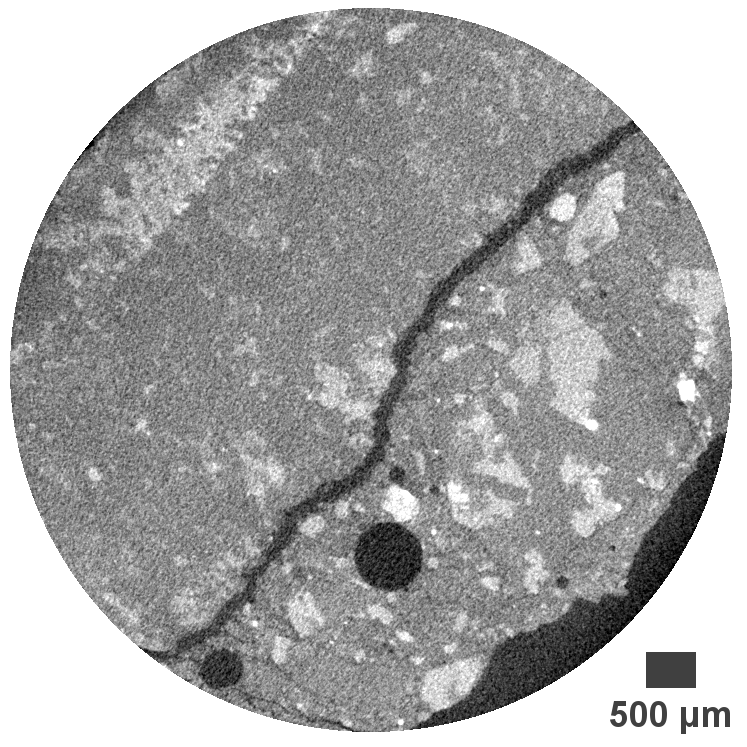}\\[-0.6mm]
(a)
\end{minipage}
\hfill
\begin{minipage}{0.49\textwidth}
\centering
\includegraphics[width=\linewidth]{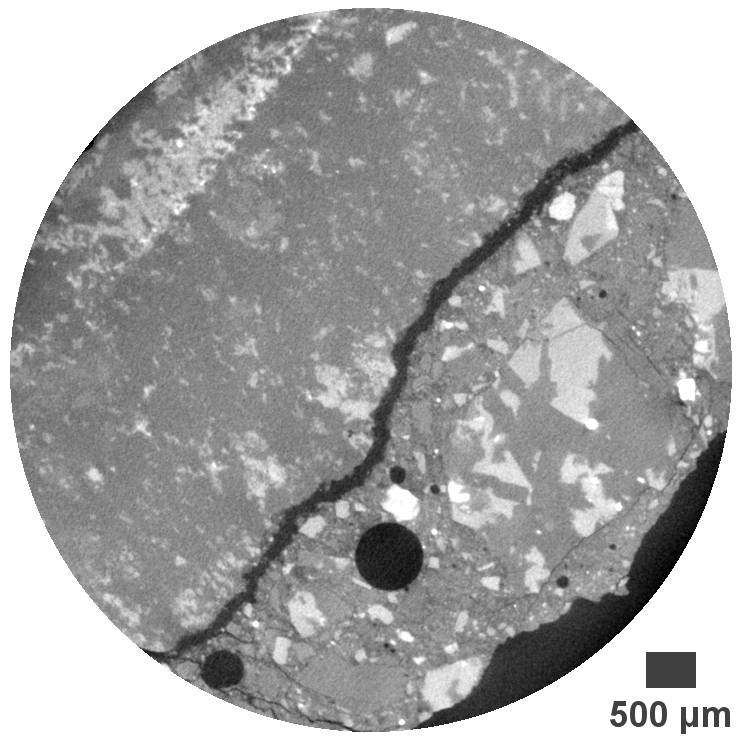}\\[-0.6mm]
(b)
\end{minipage}

\caption{Cross-sectional view of the sample: (a) fast acquisition scan, and (b) long acquisition reference scan. 
Dark pixels represent low density regions such as air-filled voids and cracks. Medium gray pixels correspond to the cement paste. Light gray regions indicate aggregate particles, while the brightest white areas represent denser mineral inclusions.}
\label{fig:sample_visualization}

\end{figure}

\subsubsection{Independent external validation dataset}
Independent external validation data were obtained from dynamic \textit{in situ} X-ray micro-CT experiments conducted during uniaxial compression using a 5 kN Deben uniaxial compression loadcell installed within the TESCAN UniTOM XL scanner. The materials consisted of recycled demolition fines composed of fine aggregates with residual hydrated cement phases attached. The material was prepared by loosely packing the fines inside a cylindrical Delrin holder placed inside the loadcell to contain the fine material during loading, as shown in Figure~\ref{fig:experimental_setup}. The specimen was positioned within the Deben stage, enabling controlled uniaxial compression while remaining within the imaging field of view.

A schematic illustration of the experimental configuration is shown in Figure~\ref{fig:experimental_setup}, including specimen placement within the loading stage and the imaging geometry. The left and right panels show slices from static scans acquired before and after compression, illustrating the internal structure of the specimen, while the center panel shows the specimen positioned within the Deben mechanical testing stage during loading. Static reference scans were acquired before and after compression using longer acquisition times and higher numbers of projections, providing high-fidelity representations of the initial and compressed states of the material.

\begin{figure}[ht]
\centering
\includegraphics[width=1\textwidth]{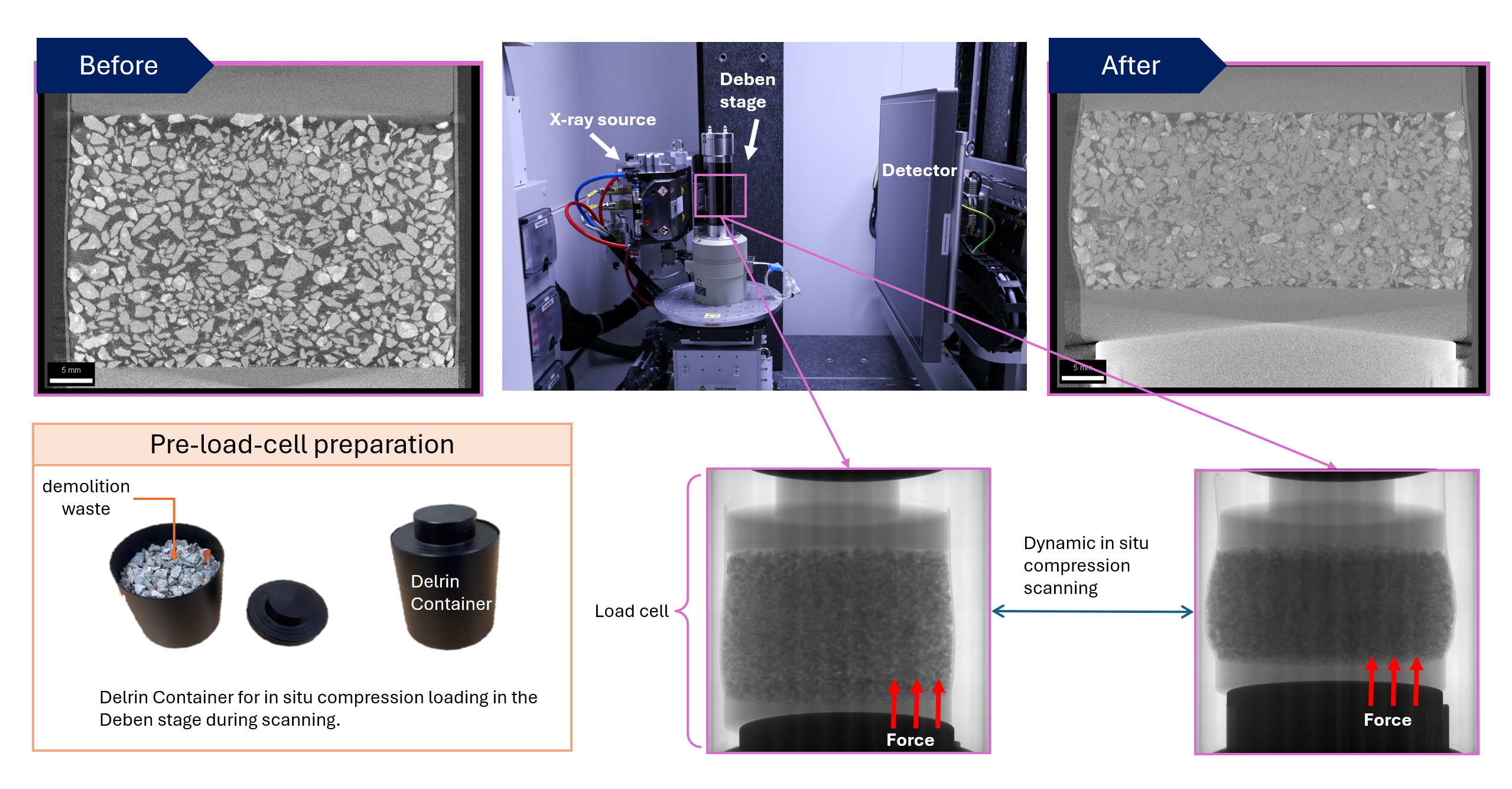}
\caption{
Experimental configuration for dynamic \textit{in situ} micro-CT imaging. 
The specimen is mounted within a Deben mechanical testing stage inside the X-ray micro-CT system.
}
\label{fig:experimental_setup}
\end{figure}

Dynamic scans were acquired during loading using reduced numbers of projections and shortened acquisition times in order to capture microstructural evolution. The static scans were reconstructed using 2856 projections, whereas the dynamic scans were acquired using 714 projections. These acquisition conditions introduce elevated noise levels and reduced spatial resolution relative to the static scans. In this work, the dynamic scans are treated as an unseen dataset and are used exclusively to evaluate the performance and generalization capability of InSituRes under realistic \textit{in situ} imaging conditions.

The static datasets serve as reference measurements, while the dynamic scans provide a challenging test case for assessing the ability of the proposed framework to recover structural detail and enable quantitative analysis under fast acquisition constraints.

\subsection{Dataset construction and preprocessing}
Training datasets were constructed from paired fast acquisition and long acquisition micro-CT scans of the same specimen. Each scan consisted of a reconstructed stack of 2D grayscale slices representing axial cross-sections of the materials scanned. The slices were read from the computer disk, sorted according to their spatial order, and assembled into 3D volumes by stacking them along the axial direction. All slices were converted to single channel floating-point images prior to further processing. To ensure spatial correspondence between the paired datasets, the fast acquisition and long acquisition volumes were required to share the same axial depth. When their in-plane dimensions differed, both volumes were center cropped to a common field of view so that paired regions had matching spatial dimensions and overlapping fields of view. Intensity normalization was applied to stabilize network training by placing each paired fast acquisition input volume and its corresponding long acquisition reference volume on a common intensity scale. Specifically, the global mean and standard deviation of the long acquisition reference volume were computed and used to perform $z$-score normalization on both the fast acquisition inputs and the corresponding reference images. 
Training samples were generated by extracting paired 3D subvolumes from the fast and long acquisition datasets. Each training sample consisted of contiguous slices from the fast acquisition volume together with the corresponding spatial region from the long acquisition reference volume. In the experiments reported here, subvolumes of size $16 \times 192 \times 192$ voxels were sampled during training. Random spatial sampling ensured the network observed a wide range of microstructural features and patterns throughout each specimen. To improve generalization and robustness to imaging variability, several data augmentation strategies were applied during training. Random in-plane rotations were applied in multiples of $90^\circ$, preserving voxel grid alignment while increasing orientation diversity. In addition, additive Gaussian noise was applied to the fast acquisition inputs, while mild intensity/contrast perturbations were applied consistently to the paired fast and long acquisition patches to simulate variations commonly observed in rapid micro-CT scanning conditions.

\subsection{Problem formulation}
Let $\mathbf{x} \in \mathbb{R}^{D \times H \times W}$ denote a X-ray micro-CT volume acquired using fast acquisition and $\mathbf{y} \in \mathbb{R}^{D \times H \times W}$ the corresponding high-quality reference volume obtained from a longer acquisition of the same specimen. Fast micro-CT scans typically employ reduced projection counts or shorter exposure times to achieve higher temporal resolution for dynamic experiments. These acquisition conditions introduce substantial degradation and artifacts that obscure fine microstructural features.

The framework is designed to recover an enhanced representation of the fast acquisition scan that approximates the structural fidelity of the long acquisition ground truth (GT) reference while remaining consistent with the measured imaging data. The enhancement network is denoted by $G_{\theta}$, parameterized by learnable parameter $\theta$. Given an input fast acquisition volume $\mathbf{x}$, the network predicts an enhanced reconstruction

\[
\hat{\mathbf{y}} = G_{\theta}(\mathbf{x}),
\]

where $\hat{\mathbf{y}}$ represents the enhanced estimate of the high-quality volume.

In conventional supervised restoration approaches, the network parameters are optimized solely by minimizing a reconstruction loss between $\hat{\mathbf{y}}$ and the reference $\mathbf{y}$. However, purely data driven training can lead to reconstructions that are visually plausible but inconsistent with the underlying imaging observations. In particular, the model may generate structures that reduce pixel-wise error relative to the reference scan but cannot explain the measured fast acquisition data.

To address this limitation, a learnable forward degradation model $A_{\phi}$ is introduced that approximates the dominant processes that degrade image quality in fast scans. The forward model maps a reconstructed high-quality estimate back to the fast acquisition domain,

\[
\tilde{\mathbf{x}} = A_{\phi}(\hat{\mathbf{y}}),
\]

where $\tilde{\mathbf{x}}$ denotes the predicted fast acquisition observation. The operator $A_{\phi}$ models key degradation mechanisms observed in rapid scans, including spatial blurring represented by a learnable point spread function, affine intensity scaling between acquisition settings, and signal-dependent noise.

To ensure numerical stability and enforce non-negativity of the attenuation-like quantities used by the forward model, both the reconstructed volume and the observed fast scan were mapped to a positive domain using a softplus transformation,

\[
\mu_{\mathrm{HR}} = \mathrm{softplus}(\hat{\mathbf{y}}), \qquad
\mu_{\mathrm{fast}}^{\mathrm{obs}} = \mathrm{softplus}(\mathbf{x}).
\]

The forward model then generates a predicted fast acquisition observation

\[
\mu_{\mathrm{fast}}^{\mathrm{pred}} = A_{\phi}(\mu_{\mathrm{HR}}).
\]

The reconstruction network and forward model are trained jointly so that the enhanced reconstruction simultaneously approximates the high-quality reference and remains consistent with the measured fast acquisition observation. The resulting optimization problem can be expressed as

\[
\min_{\theta,\phi}
\;
\mathcal{L}_{\mathrm{recon}}(\hat{\mathbf{y}},\mathbf{y})
+
\lambda_{\mathrm{phys}}
\mathcal{L}_{\mathrm{phys}}(\mu_{\mathrm{fast}}^{\mathrm{pred}},\mu_{\mathrm{fast}}^{\mathrm{obs}}),
\]

where $\mathcal{L}_{\mathrm{recon}}$ enforces similarity between the reconstructed and reference volumes and $\mathcal{L}_{\mathrm{phys}}$ penalizes discrepancies between the predicted and observed fast acquisition signals. This formulation couples volumetric image enhancement with a physics-informed observation model, thereby constraining the reconstruction process using both supervised data fidelity and physically motivated consistency.

\subsection{InSituRes network architecture}
Let $\mathbf{x}\in\mathbb{R}^{B\times 1\times D\times H\times W}$ denote a batch of fast acquisition image subvolumes, where $B$ is the batch size and $(D,H,W)$ are the volume depth and in-plane spatial dimensions. The enhancement network $G_{\theta}$ first applies a $3\times 3\times 3$ convolution to map the single channel input into a 64-channel feature space. These features are then processed by a shallow residual 3D convolutional trunk composed of two additional $3\times 3\times 3$ convolutions with an intermediate rectified linear activation. This stage provides local volumetric modeling and allows the network to refine the low-level representation without discarding the observed structure in the input scan.

The central architectural component is a slice-wise transformer inserted between the 3D convolutional encoder and the final refinement head. Rather than applying self-attention over the full 3D volume, which would be computationally prohibitive for the present task, the feature tensor is rearranged so that each axial slice is treated as an independent 2D feature map. Each slice is embedded into non-overlapping patches using a convolutional patch projection with kernel size and stride equal to 8, producing a lower-resolution token lattice with an embedding dimension of 192. Fixed 2D sinusoidal positional encodings are then added to preserve spatial identity after patchification. The patched slices are processed by six transformer blocks. Each block consists of layer normalization, multi-head self-attention with 8 attention heads, a second layer normalization, and a feed-forward multilayer perceptron with GELU activation and expansion ratio 4.0. To control memory use while retaining meaningful contextual modeling, self-attention is restricted to non-overlapping local windows of size $8\times 8$ in the embedded feature domain. This design allows the model to capture broader in-plane context than a purely convolutional network, while avoiding the computational cost of global attention across all voxels. As a result, the architecture combines local 3D feature extraction with non-local 2D contextual reasoning, which is well aligned with the anisotropic structure of reconstructed micro-CT data.

After transformer processing, the embedded features are projected back to the original in-plane voxel resolution using a transposed convolution with kernel size and stride equal to the patch size. The resulting slice-wise features are reassembled into a 3D feature volume and passed through a refinement head comprising a $3\times 3\times 3$ convolution, rectified linear activation, and a final $3\times 3\times 3$ convolution that maps the features back to a single channel residual image. The last convolution in this refinement head was initialized with zero weights and zero bias, so the network initially approximates an identity mapping. The final enhanced reconstruction is obtained by residual addition, $\hat{\mathbf{y}}=\mathbf{x}+\mathbf{r}$,
where $\mathbf{r}$ denotes the learned correction. This residual formulation is important because it biases the network toward preserving the measured structure and learning only the degradation correction required to recover a cleaner volume.

The reconstruction network was coupled to a learnable forward degradation model to enforce consistency between the enhanced reconstruction and the measured fast acquisition observation during training. To constrain the reconstruction using the imaging process itself, a learnable forward operator $A_{\phi}$ was introduced that maps the enhanced reconstruction back to the fast acquisition domain. The role of this operator is to approximate the dominant degradations that separate long acquisition and fast acquisition micro-CT scans to prevent the network from producing visually plausible but physically unsupported reconstructions.

Because the forward model is intended to represent attenuation-like image formation, the reconstructed output and the observed fast scan are first mapped to a non-negative domain using the softplus transformation,
\[
\mu_{\mathrm{HR}}=\mathrm{softplus}(\hat{\mathbf{y}}),
\qquad
\mu_{\mathrm{fast}}^{\mathrm{obs}}=\mathrm{softplus}(\mathbf{x}).
\]
The forward operator then produces a simulated fast acquisition observation,
\[
\mu_{\mathrm{fast}}^{\mathrm{pred}}=A_{\phi}(\mu_{\mathrm{HR}}).
\]

The first component of $A_{\phi}$ is a learnable two-dimensional (2D) point spread function applied independently to each axial slice. The point spread function is parameterized as an $11\times 11$ kernel initialized from a Gaussian profile with initial standard deviation 1.2. To preserve physical plausibility, the kernel coefficients are constrained to be non-negative and normalized to a unit sum. In implementation, the kernel parameters are stored in unconstrained form and transformed through a softplus nonlinearity followed by normalization. This component models the effective spatial blur associated with fast acquisition and image reconstruction.

After convolution with the learned point spread function, the blurred image is mapped into the fast-scan intensity domain using a learnable gain and bias,
\[
\mu_{\mathrm{fast}}^{\mathrm{pred}} = g\left(\mathbf{k} * \mu_{\mathrm{HR}}\right)+b.
\]
Here $\mathbf{k}$ denotes the learned point spread function, $g$ is the effective gain, and $b$ is the effective bias. In implementation, the gain is constrained to remain positive through an exponential parameterization, whereas the bias is bounded by a hyperbolic tangent parameterization scaled by a fixed maximum magnitude. These constraints stabilize optimization and limit implausible intensity drift in the forward model.

Noise in the fast acquisition data is modeled using a heteroscedastic Gaussian likelihood with signal-dependent standard deviation
\[
\sigma = \sigma_0 + \sigma_1 \sqrt{\left|\mu_{\mathrm{fast}}^{\mathrm{pred}}\right|},
\]
where $\sigma_0$ and $\sigma_1$ are learned positive parameters. This form reflects the fact that noise in X-ray micro-CT imaging is not well described by a signal-independent variance, particularly when acquisition conditions vary substantially between fast and long scans. The corresponding negative log-likelihood penalizes discrepancies between the predicted and observed fast acquisition signals while accounting for signal-dependent uncertainty.

A second consistency term is imposed in the exponential domain,
\[
\mathcal{L}_{\mathrm{beer}}
=
\frac{1}{N}\sum_i
\left|
\exp\!\left(-\mu^{\mathrm{pred}}_{\mathrm{fast},i}\right)
-
\exp\!\left(-\mu^{\mathrm{obs}}_{\mathrm{fast},i}\right)
\right|,
\]
which acts as a Beer-Lambert \cite{Swinehart1962BeerLambert, doi:10.1137/1.9780898719277} inspired constraint. This term is not intended as a full physical model of tomographic projection formation, since the data are already reconstructed image volumes, but instead as an additional regularizer that promotes consistency in an attenuation-related intensity domain.

The forward model is further regularized using a spatial smoothness penalty on the learned point spread function, together with weak anchoring terms that keep the effective gain close to unity and the effective bias close to zero. These regularizers reduce degeneracy in the joint optimization and help the forward operator remain interpretable as a degradation model rather than an unconstrained auxiliary network.

\subsection{Training procedure}
InSituRes was implemented in PyTorch \cite{Paszke2019PyTorch}. The enhancement network $G_{\theta}$ and the forward degradation model $A_{\phi}$ were trained jointly using paired fast acquisition and long acquisition micro-CT volumes of the same specimen. Training samples were extracted as 3D slabs from spatially corresponding regions of the paired datasets. Each slab contained $16\times192\times192$ voxels (depth $\times$ height $\times$ width). Intensities were normalized using global $z$-score statistics derived from the long acquisition reference volume.

Model parameters were optimized using the AdamW optimizer \cite{loshchilov2019decoupled} with cosine learning rate scheduling \cite{loshchilov2017sgdr}. Training was performed using mixed-precision arithmetic on an NVIDIA GeForce RTX 5090 GPU.

The objective function combined reconstruction fidelity, structural similarity, and physics-based observation consistency. The reconstruction term encouraged agreement between the enhanced reconstruction $\hat{\mathbf{y}}$ and the long acquisition reference $\mathbf{y}$ using a robust voxel-wise loss together with a structural similarity term computed slice-wise in the in-plane dimensions. To incorporate imaging physics into the reconstruction process, additional loss terms were introduced that enforce consistency between the simulated fast acquisition observation produced by the forward model and the measured fast scan. These physics-based terms include a likelihood-based discrepancy in the attenuation domain and a consistency constraint in the exponential intensity domain inspired by the Beer -Lambert relationship.

The overall objective can therefore be expressed as

\[
\mathcal{L} =
\lambda_{\mathrm{rec}}\mathcal{L}_{\mathrm{rec}}
+
\lambda_{\mathrm{phys}}\mathcal{L}_{\mathrm{phys}},
\]

where $\mathcal{L}_{\mathrm{rec}}$ represents the reconstruction fidelity terms and $\mathcal{L}_{\mathrm{phys}}$ enforces compatibility between the reconstructed volume and the observed fast acquisition data through the forward degradation model.

To stabilize optimization, the influence of the physics-consistency terms was gradually increased during training so that the network first learned a stable reconstruction before stronger physical constraints were imposed. Training was initially performed using randomly sampled spatial patches to encourage robust feature learning across the specimen. In the final stage of training, the model was fine-tuned on full in-plane slices while maintaining the same slab depth, improving compatibility between patch-based training and full-volume inference.

\subsection{Conventional filtering baseline comparison}
Conventional post-processing baselines were evaluated in Fiji \cite{Schindelin2012Fiji} to compare InSituRes with standard image filtering approaches applied to the fast acquisition scans. Filters were applied to the representative fast acquisition slice shown in Figure~\ref{fig:reconstruction_comparison}, which was also used for the corresponding long acquisition reference and InSituRes-enhanced comparison. The evaluated baselines included Gaussian filtering, median filtering, bilateral filtering \cite{Zhang2008MultiresolutionBilateral}, and anisotropic diffusion \cite{56205}. Gaussian filtering was evaluated using Fiji Gaussian blur sigma values of 1, 2, and 3 pixels. Median filtering was evaluated using Fiji median filter radii of 1, 2, and 3 pixels. Bilateral filtering was evaluated using spatial radius values of 2, 3, and 4 pixels with corresponding intensity domain parameters of 15, 25, and 35. Anisotropic diffusion was evaluated using 5, 10, and 20 iterations. For each filtering family, the parameter setting that produced the lowest absolute porosity error relative to the long acquisition reference scan was selected for reporting in the representative slice baseline comparison. After filtering, each image was segmented using the same unsupervised clustering procedure applied to the fast acquisition, long acquisition reference, and InSituRes-enhanced images. Porosity was calculated as the percentage of pixels assigned to the pore phase. Absolute porosity error was calculated relative to the long acquisition reference scan.

\subsection{Inference on full volumes}
During inference, the trained reconstruction network was applied directly to fast acquisition micro-CT volumes. Because the network operates on fixed depth slabs, each input volume was divided into contiguous axial segments of depth 16 and processed sequentially. Each slab was normalized using the stored training statistics, passed through the network, and transformed back to the original intensity scale after reconstruction. The slab-wise outputs were then concatenated along the axial direction to recover the full enhanced volume. The exponential moving average parameters of the reconstruction network were used at inference when available. No explicit overlap or blending between neighboring slabs was applied in the present implementation. An optional global post-contrast adjustment around the volume mean was implemented but set to unity by default.

\subsection{Evaluation metrics and statistical analysis}

Model performance was evaluated by comparing enhanced reconstructions with their corresponding long acquisition reference volumes. Peak signal-to-noise ratio (PSNR) was used to quantify voxel-wise reconstruction fidelity,
\[
\mathrm{PSNR} = 20\log_{10}(R) - 10\log_{10}(\mathrm{MSE}),
\]
where $R$ denotes the dynamic range of the reference volume and $\mathrm{MSE}$ is the mean squared error between the enhanced and reference reconstructions.

Structural similarity was quantified using SSIM computed independently on each axial slice with an $11\times 11$ Gaussian weighting window of standard deviation 1.5 and then averaged across slices to obtain a volume-level score. This slice-wise evaluation reflects the principal perceptual degradations in fast scans are manifested in the in-plane image structure.

The full depth of each paired volume was partitioned along the axial direction into training, validation, and test regions using a 70:15:15 split. Validation performance was monitored over fixed samples drawn from the validation region, and the best-performing checkpoint was selected using mean validation PSNR. Reported quantitative results were obtained by averaging metrics across the evaluation samples and reporting the mean values for PSNR and SSIM.

\end{flushleft}

\section*{Funding}
\begin{flushleft}
The Duke University Department of Civil and Environmental Engineering is greatly acknowledged for their support. 
\end{flushleft}

\section*{Acknowledgments}
\begin{flushleft}
The authors thank Anay Tillu for assistance with preliminary data inspection and manual identification of misaligned images in an initial subset of the dataset during early stage testing. This contribution was limited to exploratory analysis of a partial dataset and was not used in the final workflow, which was conducted on the full dataset without manual selection. 
The Duke University Department of Civil and Environmental Engineering is also greatly acknowledged for their support. 
\end{flushleft}

\section*{Data availability}
\begin{flushleft}
The X-ray micro-CT datasets generated and analyzed during this study are not publicly available at the time of submission. The data will be made available to editors and reviewers upon reasonable request during peer review. The data will be deposited in an appropriate public repository before publication, subject to repository size limits and institutional data-sharing requirements.
\end{flushleft}

\section*{Code availability}
\begin{flushleft}
At the time of submission, the code used to train and evaluate InSituRes is not yet publicly available in an open source repository. However, the code will be made available privately to editors and reviewers upon request during peer review. The code will be deposited in an appropriate public repository shortly after submission and no later than publication.
\end{flushleft}

\section*{Declaration of competing interest}
The authors declare that they have no known competing financial interests or personal relationships that could have appeared to influence the work reported in this paper.

\section*{Author Contribution Declaration}

Q.T. conceptualized the study, developed the methodology, performed software related work, validated the results, conducted formal analysis and investigation, curated the data, prepared the visualizations, and wrote the original draft. A.B. contributed to validation, investigation, resources, and review \& editing of the manuscript. S.T. contributed to resources, data curation, and review \& editing of the manuscript. S.C. contributed to investigation, review \& editing of the manuscript. L.E.D. contributed to conceptualization, formal analysis, resources, data curation, visualization, supervision, project administration, funding acquisition, and review \& editing of the manuscript.

\bibliographystyle{naturemag}
\bibliography{030926ref}


\end{document}